\documentclass[conference,compsoc]{IEEEtran}
\usepackage[table,dvipsnames]{xcolor}
\usepackage{graphicx}
\usepackage{amsmath,amssymb}
\usepackage{xstring,xspace,xparse,xfp}
\usepackage{siunitx}
\usepackage{enumitem}
\usepackage{booktabs,tabularx,array,multirow,makecell}
\usepackage{caption} % Restore the previous 10 pt figure/table caption style.
\usepackage{float} % Existing tables use [H].
\usepackage{balance}
\usepackage{tikz}
\usepackage{algorithm,algpseudocode}
\usepackage{pifont}
\usepackage{listings}
\usepackage{tcolorbox}
\tcbuselibrary{skins,breakable,listings}

\newcommand{\circlednum}[1]{%
  \tikz[baseline=(char.base)]
  \node[shape=circle,fill=black,text=white,
        inner sep=1.2pt,font=\scriptsize] (char) {#1};%
}

\definecolor{themeBlue}{HTML}{305F82}
\definecolor{themeGreen}{HTML}{316653}
\definecolor{themeAmber}{HTML}{896519}
\definecolor{themeGray}{HTML}{59616B}
\definecolor{themeYellow}{HTML}{B59A3D}
\definecolor{themeOrange}{HTML}{BF7842}
\definecolor{backgroundColor}{HTML}{F8F9FB}

\newcommand{\accdigits}{1}
\DeclareRobustCommand{\formatdecimal}[1]{%
  \num[mode=text,detect-weight=true,detect-family=true,
    round-mode=places,round-precision=\accdigits,
    round-pad=true,round-half=up]{#1}}
\DeclareRobustCommand{\acc}[1]{%
  \begingroup
  \def\accuracyvalue{#1}%
  \ifdim#1pt<100pt
    \ifdim\fpeval{round(#1,\accdigits)}pt=100pt
      \edef\accuracyvalue{\fpeval{100-10^(-\accdigits)}}%
    \fi
  \fi
  \expandafter\formatdecimal\expandafter{\accuracyvalue}%
  \endgroup}
\DeclareRobustCommand{\accp}[1]{\acc{#1}\%}

\definecolor{accBelow70}{HTML}{FFE08A}
\definecolor{acc70}{HTML}{FFEBAA}
\definecolor{acc80}{HTML}{FFF4CC}
\definecolor{acc90}{HTML}{F2F8EE}
\definecolor{acc95}{HTML}{EAF5E5}
\definecolor{acc99}{HTML}{DEEFDA}
\definecolor{acc100}{HTML}{D1E9C9}
\newcommand{\acccell}[1]{%
  \ifdim#1pt<70pt\cellcolor{accBelow70}%
  \else\ifdim#1pt<80pt\cellcolor{acc70}%
  \else\ifdim#1pt<90pt\cellcolor{acc80}%
  \else\ifdim#1pt<95pt\cellcolor{acc90}%
  \else\ifdim#1pt<99pt\cellcolor{acc95}%
  \else\ifdim#1pt<100pt\cellcolor{acc99}%
  \else\cellcolor{acc100}%
  \fi\fi\fi\fi\fi\fi\textcolor{black}{\acc{#1}}}

\newif\ifdraftnotes
\draftnotestrue

\newcommand{\bheading}[1]{\par\addvspace{3pt}\noindent\textbf{#1}\enspace\ignorespaces}
\newcommand{\tabref}[1]{Table~\ref{#1}}

\setlist[itemize]{leftmargin=12pt,labelwidth=0pt,labelsep=0.5em,
  align=left,itemsep=0pt,topsep=2pt,parsep=0pt,partopsep=0pt}
\newenvironment{packeditemize}{
    \begin{list}{$\bullet$}{
        \setlength{\labelwidth}{4pt}
        \setlength{\itemsep}{0pt}
        \setlength{\leftmargin}{\labelwidth}
        \addtolength{\leftmargin}{\labelsep}
        \setlength{\parindent}{0pt}
        \setlength{\listparindent}{\parindent}
        \setlength{\parsep}{0pt}
        \setlength{\topsep}{1pt}
    }
}{
    \end{list}
}

\tcbset{paper/base/.style={enhanced,breakable,colback=backgroundColor,
  colframe=themeGray,colbacktitle=themeGray!8,coltitle=black,
  boxrule=0.45pt,arc=1pt,left=5pt,right=5pt,top=4pt,bottom=4pt,
  boxsep=1pt,before skip=6pt,after skip=6pt,fonttitle=\small\bfseries,
  fontupper=\small}}

\tcbset{paper/rail/.style={paper/base,frame hidden,boxrule=0pt,
  sharp corners,colback=black!2,
  left=7pt,right=5pt}}
\tcbset{paper/statement/.style={paper/rail,colback=black!4,
  fontupper=\normalsize,before skip=4pt,after skip=6pt,
  top=4pt,bottom=4pt,boxsep=0pt}}
\newtcolorbox[auto counter]{challengebox}[1][]{paper/statement,
  borderline west={1.6pt}{0pt}{themeOrange},
  before upper={\textbf{Challenge~\thetcbcounter.}\enspace},#1}
\newcommand{\challenge}[2][]{%
  \begin{challengebox}[#1]#2\end{challengebox}}
\newtcolorbox[auto counter]{findingbox}[1][]{paper/statement,
  borderline west={1.6pt}{0pt}{themeBlue},
  before upper={\textbf{Finding~\thetcbcounter.}\enspace},#1}
\newcommand{\finding}[2][]{%
  \begin{findingbox}[#1]#2\end{findingbox}}

\newcommand{\downchange}[1]{\textcolor[RGB]{150,0,0}{\scriptsize\textbf{\ensuremath{\downarrow}#1}}}

\providecommand{\evalmodel}[2]{%
  \begingroup
  \def\evalmodelicon{#1}%
  \def\evalmodelheight{1em}%
  \IfStrEq{#1}{kimi}{\def\evalmodelicon{moonshot}\def\evalmodelheight{0.85em}}{}%
  \IfStrEq{#1}{moonshot}{\def\evalmodelheight{0.85em}}{}%
  \makebox[1em][c]{\raisebox{-0.2ex}{%
    \includegraphics[height=\evalmodelheight]{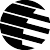}}}%
  \hspace{0.35em}#2%
  \endgroup
}

\newcommand{\agentIcon}[2]{%
  \makebox[1em][c]{\raisebox{-0.2ex}{%
    #1}}%
  \hspace{0.35em}#2%
}

\ifCLASSOPTIONcompsoc
  \usepackage[nocompress]{cite}
\else
  \usepackage{cite}
\fi
\usepackage{xurl}
\usepackage[hidelinks]{hyperref}

\begin{document}
\title{CAPTA: Identifying LLMs Behind AI Agents}
\author{\IEEEauthorblockN{Kailun Yan, Yue Huang, and Sisi Duan}
\IEEEauthorblockA{Tsinghua University\\
\href{mailto:kailun@mail.tsinghua.edu.cn}{\nolinkurl{kailun@mail.tsinghua.edu.cn}},
\href{mailto:yuehuang2021@outlook.com}{\nolinkurl{yuehuang2021@outlook.com}},
\href{mailto:duansisi@tsinghua.edu.cn}{\nolinkurl{duansisi@tsinghua.edu.cn}}}}
\maketitle

\begin{abstract}
Large language model (LLM)-based agents access remote services through tools to complete user tasks, often via API calls. In this setting, a remote service may want to identify the backend LLM of an agent to launch model-specific attacks, including prompt injection or selective delivery of manipulated content. However, the service cannot directly query the backend model and must instead collect observable responses through its interactions with the agent, making identification challenging.

In this paper, we present Completely Automated Preference Test to tell LLMs Apart (CAPTA), a model identification approach inspired by CAPTCHA (Completely Automated Public Turing test to tell Computers and Humans Apart), which distinguishes humans from automated programs. CAPTA presents multiple-choice questions as verification challenges, using differences in models' selection preferences to identify the backend LLM. It selects questions based on cross-model option-generation diversity and uses LLM-guided search to optimize their option sets. Our evaluation shows that CAPTA achieves 99.0\% mean identification accuracy with only 10 questions across 25 LLMs from 10 model families. We further evaluate CAPTA with four widely used agents and seven representative LLMs, achieving 99.7\% mean identification accuracy across all 28 agent--LLM combinations with 30 submitted answers.

\end{abstract}

% CAPTA Story
% An agent accesses a remote service.
% But what is the LLM behind this agent?
% Why does model identity matter?
% Can the service turn verification into an identification opportunity?
% Our solution: CAPTA, inspired by CAPTCHA,
% Different LLMs exhibit different preferences.
% CAPTA makes this idea practical.

\section{Introduction}
\label{sec:intro}

Large language models (LLMs) have advanced rapidly, enabling LLM-based agents to perform complex tasks autonomously~\cite{317937, DBLP:conf/nips/DebenedettiZBB024}.
An important capability of these agents is to use tools to retrieve information from external websites, APIs, and other remote services~\cite{DBLP:conf/uss/KimSN0L25}.
The scale of such interactions is growing rapidly. 
According to DataDome~\cite{datadome2026aitraffic}, AI agent traffic exceeded 6.6 billion requests in a single month in Q2 2026. 
Another study found that agents made an average of 3.64 external requests per session when performing information retrieval tasks~\cite{DBLP:conf/sigir/NingCKLMMCX26}.

A remote service may want to identify the backend LLM of an agent.
This information may have security implications because different LLMs can exhibit different vulnerabilities and safety behaviors~\cite{DBLP:conf/nips/ChaoDRACSDFPTH024,307744}.
Knowing the backend model may enable model-specific attacks. 
For example, the remote service may select model-specific prompt injection content~\cite{DBLP:conf/iclr/LiaoMXKZXTL025}, as illustrated in Anthropic's analysis of browser-agent security~\cite{anthropic2025promptinjectiondefenses}.
More broadly, a service could selectively deliver manipulated content to AI agents while showing the original content to human visitors. This technique is known as \textit{cloaking}~\cite{acharya2021phishprint, DBLP:conf/sp/LiQJWGJWD26, DBLP:conf/uss/AyzenshteynWM25}.
For example, shopping websites could use AI-targeted cloaking to show AI agents fake promotions or product descriptions while showing original content to human visitors, influencing agents' recommendations or purchasing decisions~\cite{splx2025aitargetedcloaking}.

\begin{figure}[!t]
  \centering
  \includegraphics[width=\linewidth]{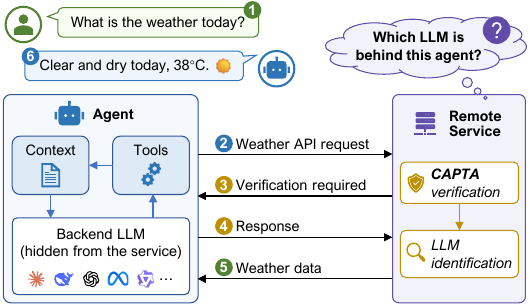}
    \caption{An LLM-based agent retrieves weather information from a remote service. The agent's backend LLM is unknown to the service.}
  \label{fig:capta}
\end{figure}

As shown in Figure~\ref{fig:capta}, consider a common scenario where a user asks an agent to check today's weather \circlednum{1}. 
The agent sends the request to its backend LLM, which decides to call a tool to access an external weather API \circlednum{2}. 
Normally, the weather service directly returns the requested data \circlednum{5}, and the agent summarizes the results for the user \circlednum{6}.
Meanwhile, some remote services may require CAPTCHA~\cite{vonahn2003captcha}, a challenge designed to distinguish humans from bots, before returning the data \circlednum{3}.
However, modern LLM-based agents can solve many of these challenges and may actively attempt~\cite{DBLP:conf/uss/Teoh0LLSH025, DBLP:conf/ccs/DengO000025, DBLP:conf/sp/AliKICKP27} to do so to complete their tasks \circlednum{4}.
% \sisi{add citation and maybe briefly describe what it is?} 

In such a context, identifying the backend LLM behind an agent is challenging, mainly because the remote service does not directly interact with the LLM behind the agent. 
Namely, any interactions beyond the intended task may be treated as malicious attempts and rejected by the agent.
Therefore, although some existing works attempt to identify LLMs under different settings, they either use designed inputs to elicit and analyze responses from the backend LLM~\cite{DBLP:conf/uss/PasquiniKA25, DBLP:conf/acl/GubriULYO24, DBLP:conf/acl/HuJLAHHG26, tsai2026rafpidentifyingllmlineages, 10735575}, analyze agent behaviors~\cite{DBLP:journals/corr/abs-2609-28559, DBLP:journals/corr/abs-2609-00052}, or detect pre-embedded watermarks in model outputs~\cite{DBLP:conf/nips/NaseryHBSTVO25, DBLP:conf/acl/YamabeW0W25, DBLP:conf/aaai/XiongWYMYPDW26, DBLP:conf/iclr/GloaguenSJV26}.
Thus, existing approaches have limited applicability to the settings considered in this work.

%\sisi{what does it mean?}

\bheading{Our work.}
Inspired by CAPTCHA, we propose CAPTA (Completely Automated Preference Test to tell LLMs Apart), a new approach to identifying the backend LLM behind an agent. 
When an agent accesses a remote service, it may encounter a CAPTCHA challenge and try to solve it to continue its task.
The remote service uses this opportunity to replace traditional CAPTCHA challenges with carefully designed multiple-choice questions, called probes.
The agent answers these questions during its normal interactions with the remote service, and the service uses the responses to identify the agent's backend LLM.
Our work focuses on the problem of identifying backend LLMs and does not consider subsequent attacks against the identified models.

CAPTA is based on the observation that different LLMs may exhibit different preferences when answering the same question. 
For example, when asked to select a rainbow color in 20 independent trials, 
DeepSeek-V4-Flash and Qwen-3.7-Flash consistently select \textit{red} (20/20), while GPT-5.6-Luna and Claude-Sonnet-5 consistently select \textit{blue} (20/20). Our motivation is thus to design the probes as multiple-choice questions to distinguish the LLMs.

To construct discriminative probes, CAPTA addresses two challenges: how to select questions that can distinguish different LLMs, and how to find effective option sets for each question. 
First, CAPTA uses multiple LLMs to generate candidate questions and options, then ranks questions based on the diversity of generated options.
Second, it uses an LLM-guided search method to optimize the option sets of selected questions, further improving their ability to distinguish different LLMs.
Combining multiple probes allows the service to identify the backend LLM.

We evaluate CAPTA on 25 well-known LLMs. With only 10 probes, CAPTA achieves a mean identification accuracy of 99\%, requiring only one or two agent--service interactions in practice. 
With 21 probes, it achieves at least 99\% accuracy for every candidate model.
The results also show that our methods effectively address both challenges, improving single-probe and joint identification accuracy.

We conduct extensive experiments to evaluate CAPTA's practical feasibility and robustness.
Evaluations with four widely used agents and seven LLMs show that all agents successfully complete CAPTA verification across 28 agent--LLM combinations.
CAPTA achieves \accp{99.7} mean identification accuracy within three agent--service interactions.
We further evaluate CAPTA across different task contexts, showing that it maintains high identification accuracy despite context-induced preference shifts.
We also evaluate its ability to detect unknown models in open-set settings.

\bheading{Contributions.}
We make the following contributions.

\begin{packeditemize}
    
    \item \textbf{LLM identification approach.}
    We propose CAPTA, a new approach that enables remote services to accurately identify the backend LLM behind an agent in just a few agent--service interactions.
    
    \item \textbf{Automated probe construction.}
    We develop an automated framework to address the two challenges in probe construction.
    We propose diversity scores to rank and select promising questions, and an LLM-guided method to optimize their option sets.
    
    \item \textbf{Extensive evaluation.}
    We evaluate CAPTA on 25 well-known LLMs, achieving 99.0\% mean identification accuracy with only 10 probes.
    We further evaluate CAPTA with four widely used agents and seven LLMs, demonstrating its practical feasibility.
    
\end{packeditemize}

\bheading{Artifact availability.}
We provide an anonymous online demo of CAPTA at \url{https://capta-demo.pages.dev/}, which includes the reference fingerprints and evaluation data. We will release the source code upon acceptance.

\section{Background and Related Work}
\label{sec:bg}

% This section introduces agent--service interactions, reviews existing approaches, and presents our threat model.

\begin{figure*}[!t]
  \centering
  \includegraphics[width=0.85\linewidth]{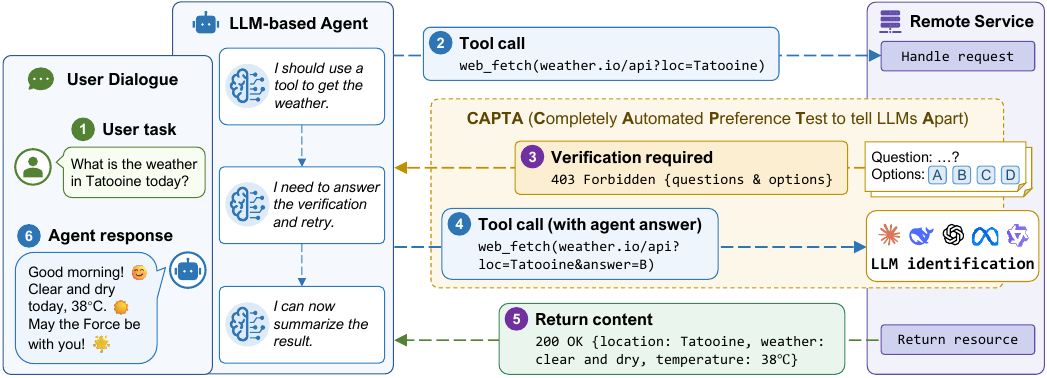}
    \caption{An LLM-based agent performing a weather query. The remote weather service can use CAPTA as a CAPTCHA verification to identify the agent's backend LLM.}
  \label{fig:example}
\end{figure*}

\subsection{Agent--Service Interaction}

LLM-based agents, such as OpenClaw~\cite{openclaw_runtime}, typically consist of a backend LLM and an execution framework~\cite{yao2023react}. 
Given a user task, the LLM decides what to do next and may invoke tools to access external resources. 
The execution framework handles these tool calls and returns their results to the LLM, which uses them to decide its next action. 
This process continues until the task is completed.

As shown in Figure~\ref{fig:example}, a user asks an agent about today's weather \circlednum{1}. 
The backend LLM decides to invoke \texttt{web\_fetch}, which sends a request to a remote weather service \circlednum{2}. 
Normally, the service returns the requested data directly \circlednum{5}. 
The LLM then summarizes the results and responds to the user \circlednum{6}. 
Similar workflows are common in agent applications, such as news retrieval and flight tracking.

\bheading{CAPTCHA verification.}
To restrict automated access and prevent excessive requests, remote services may require CAPTCHA verification before providing requested resources.
As shown in Figure~\ref{fig:example},
the service first returns a CAPTCHA challenge with \texttt{403 Forbidden} to the agent \circlednum{3}.
The agent processes the challenge and submits its response to the service \circlednum{4}.
Upon successful verification, the service returns the requested resource \circlednum{5}.

A CAPTCHA (Completely Automated Public Turing test to tell Computers and Humans Apart) is a verification mechanism originally designed to distinguish humans from automated programs~\cite{vonahn2003captcha}.
With advances in LLMs, agents can now solve some CAPTCHA challenges autonomously~\cite{DBLP:conf/ccs/DengO000025, 320599, luo2025opencaptchaworld}, allowing them to continue tasks without human intervention.
Our CAPTA leverages this verification process to identify the agent's backend LLM.
% Instead of traditional CAPTCHA challenges, CAPTA presents crafted probes and identifies the LLM based on the agent's responses.

\begin{table*}[t]
\centering
\caption{Comparison of CAPTA with existing identification and detection approaches.}
\label{tab:related-identification}
\begingroup
\footnotesize
\setlength{\tabcolsep}{2pt}
\setlength{\extrarowheight}{0pt}
\renewcommand{\arraystretch}{1.25}

\begin{tabularx}{\linewidth}{@{}
>{\hsize=0.75\hsize\raggedright\arraybackslash\hspace*{4pt}}X
 >{\hsize=1.65\hsize\raggedright\arraybackslash}X
 >{\hsize=0.95\hsize\raggedright\arraybackslash}X
 >{\hsize=0.60\hsize\centering\arraybackslash}X
 >{\hsize=1.05\hsize\centering\arraybackslash}X@{}}
\toprule
\textbf{Approach}
& \textbf{Purpose}
& \textbf{Observable Evidence}
& \textbf{Direct Access}
& \textbf{Service-Side Identification} \\
\midrule

Browser behavior
& Backend model identification~\cite{DBLP:journals/corr/abs-2605-14786}
& Browser interaction traces
& No
& Partial \\

\addlinespace[1pt]

\rowcolor{gray!8}
& Backend LLM identification~\cite{DBLP:journals/corr/abs-2609-28559}
& Execution trajectories~\cite{DBLP:journals/corr/abs-2609-28559}
&
& \\

\rowcolor{gray!8}
\multirow{-2}{*}{Agent behavior}
& Provider auditing~\cite{DBLP:journals/corr/abs-2609-00052}
& Tool-call distributions~\cite{DBLP:journals/corr/abs-2609-00052}
& \multirow{-2}{*}{Yes (Agent)}
& \multirow{-2}{*}{No} \\

\addlinespace[1pt]

& Model identification~\cite{DBLP:conf/uss/PasquiniKA25, DBLP:conf/acl/GubriULYO24, DBLP:conf/acl/HuJLAHHG26, DBLP:journals/corr/abs-2607-10252}
&
&
& \\

\multirow{-2}{*}{Generated text}
& Provenance analysis~\cite{tsai2026rafpidentifyingllmlineages, 10735575, DBLP:conf/nips/NikolicBS25, DBLP:conf/icml/SunYXK025, DBLP:journals/corr/abs-2607-25880}
& \multirow{-2}{*}{Model outputs}
& \multirow{-2}{*}{Yes (Model)}
& \multirow{-2}{*}{No} \\

\addlinespace[1pt]

\rowcolor{gray!8}
& Ownership verification~\cite{DBLP:conf/nips/NaseryHBSTVO25, DBLP:conf/acl/YamabeW0W25, DBLP:conf/aaai/XiongWYMYPDW26, DBLP:conf/iclr/GloaguenSJV26}
&
& \mbox{Yes (Model)}
& \\

\rowcolor{gray!8}
\multirow{-2}{*}{Watermarking}
& AI-generated text detection~\cite{DBLP:conf/iclr/KirchenbauerGWS24, DBLP:conf/iclr/ZhaoA0W24, DBLP:conf/sp/Cohen0S25}
& \multirow{-2}{*}{\makecell[l]{Model outputs\\(pre-embedded signals)}}
& No
& \multirow{-2}{*}{Partial} \\

\addlinespace[1pt]

\textbf{CAPTA (Ours)}
& \textbf{Backend LLM identification}
& \textbf{Agent-submitted choices}
& \textbf{No}
& \textbf{Yes} \\

\bottomrule
\end{tabularx}

\par\smallskip
\parbox{\linewidth}{\footnotesize
Direct Access means directly querying the target model or agent.
Partial indicates applicability in some settings, but not directly in ours.
}
\endgroup
\end{table*}

%Existing studies have explored different techniques for identifying LLMs, verifying model origins, and tracing model outputs.

\subsection{Existing Approaches to LLM Identification}
\label{sec:bg:existing}
Identifying underlying LLMs has attracted increasing interest for various purposes, as summarized in \tabref{tab:related-identification}.
Beyond security, existing works aim to audit agent model claims, trace model origins, and detect AI-generated text.

% \sisi{XXXX (summarize the purposes)?}
% \sisi{@Kailun: double check the revision in this section.}
% (Note: not all works focus on LLM identification; some detect AI agents.)

\bheading{Browser behavior.} As agents browse websites on users' behalf, their observable interactions provide websites with opportunities to identify the underlying LLMs. It was found that browser agents' UI actions and interaction timings can reveal their backend LLMs, potentially allowing websites to select model-specific attacks~\cite{DBLP:journals/corr/abs-2605-14786}.

\bheading{Agent behavior.} 
Identifying the models behind agentic systems can help users audit whether agents use their providers' claimed LLMs or are silently served different models.
This is often achieved by analyzing agents' execution data collected from specific tasks.
For example,~\cite{DBLP:journals/corr/abs-2609-28559} identifies LLMs through execution trajectories of coding agents, while AgentProv~\cite{DBLP:journals/corr/abs-2609-00052} analyzes tool-call distributions to audit whether agentic APIs use their claimed models.

\bheading{Generated text.}
Model responses can help determine which LLM an application uses or whether a suspect model was derived from a known source model~\cite{DBLP:conf/acl/HuJLAHHG26, DBLP:journals/corr/abs-2607-10252, DBLP:conf/nips/NikolicBS25, DBLP:conf/icml/SunYXK025}.
For example, LLMmap~\cite{DBLP:conf/uss/PasquiniKA25} sends carefully designed queries to an application and identifies its LLM based on the model outputs.
TRAP~\cite{DBLP:conf/acl/GubriULYO24} uses adversarial prompts to make a target model produce predefined outputs while other models produce different ones.
Stemma~\cite{DBLP:journals/corr/abs-2607-25880} uses selected multiple-choice questions with fixed option sets to test whether a target model shares a source model's lineage.
These works often directly query the target LLM or application and collect the model's outputs for analysis.
% \sisi{consider briefly describe what are the queries since CAPTA also uses a similar idea?}
% (note: these methods analyze model outputs, while CAPTA relies on option selections. I will clarify it in comparison.)

\bheading{Watermarking.} 
Watermarking is an extensively studied topic where model developers embed identifiable signals into their models to verify ownership through model responses~\cite{DBLP:conf/naacl/Xu0MKXC24, DBLP:conf/nips/NaseryHBSTVO25, DBLP:conf/acl/YamabeW0W25, DBLP:conf/aaai/XiongWYMYPDW26, DBLP:conf/iclr/GloaguenSJV26}.
Another approach is \emph{text watermarking}, which embeds statistical signals into generated text by modifying token selection or generation probabilities~\cite{DBLP:conf/iclr/KirchenbauerGWS24, DBLP:conf/iclr/ZhaoA0W24, DBLP:conf/sp/Cohen0S25}.
Detectors then analyze these signals to identify AI-generated text or trace its source.

\bheading{Comparison with CAPTA.}
As shown in Table~\ref{tab:related-identification}, CAPTA differs from existing approaches.
CAPTA considers a remote service that interacts with an agent through its own service interface, without direct access to the agent or its backend LLM.
Browser-based approaches focus on web scenarios, whereas CAPTA applies to broader agent--service interaction scenarios.
Agent-behavior-based and generated-text-based approaches require access to agent execution data or the target model and its outputs, which are unavailable in our setting.
Meanwhile, the service does not control model training or release and thus cannot require watermarks to be embedded in advance.

Some existing approaches also use carefully designed questions and multiple-choice tests. However, these approaches still require direct model access and do not address automated probe construction and optimization.
In contrast, CAPTA leverages the normal CAPTCHA verification to identify backend LLMs.
We further develop diversity-based question selection and LLM-guided option-set optimization to construct discriminative probes.

\subsection{Threat Model}

Based on the agent--service interactions described above,
we consider a remote service provider as the adversary.
Its goal is to identify the backend LLM used by an agent accessing its service.
Our threat model focuses exclusively on model identification;
how the identified model may subsequently be exploited is beyond our scope.

The adversary can provide remote services through various interfaces, including APIs, websites, Skills, and Plugins.
It can receive requests from agents and control the responses returned by its services.
Through these interfaces, the adversary can present CAPTA probes without accessing the agent's internal information.
It can repeat interactions across multiple service accesses to collect additional responses.
It can also access candidate LLMs through existing LLM providers and test them in advance
to characterize their behavioral preferences.

\begin{figure*}[t]
  \centering
  \includegraphics[width=\linewidth]{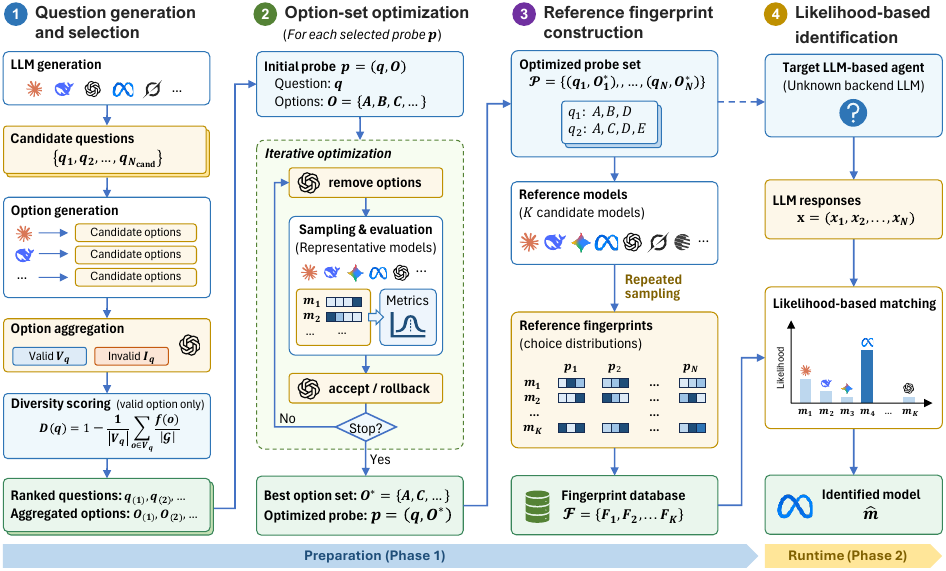}
    \caption{
        Overview of the CAPTA workflow.
        In the preparation phase, CAPTA generates and selects questions (Stage 1), 
        optimizes their options (Stage 2), 
        and records each model's choice frequencies as reference fingerprints (Stage 3).
        In the runtime phase, 
        it compares the agent's answers with these fingerprints to identify the backend LLM (Stage 4).
        }
  \label{fig:design-overview}
\end{figure*}

\section{CAPTA Overview}
\label{sec:ov}
% This section introduces the rationale behind probe design, its two challenges, and the CAPTA workflow.

\subsection{Probe Design and Challenges}

\bheading{Preference instead of capability.}
Existing LLM benchmarks~\cite{DBLP:conf/nips/BeanKRHMBESKBDB25, DBLP:conf/naacl/KimSCLKYSCSBPHJCSLOLHJK25} mainly evaluate model capabilities through tasks such as mathematical reasoning and code generation.
However, these tests are not suitable for model identification, as different LLMs may have similar capabilities and produce the same correct answers.
In contrast, different LLMs may exhibit different preferences when answering the same question.
For example, when asked to recommend an Asian dish, Claude-Sonnet-5 may prefer ramen, while DeepSeek-V4-Flash may choose rice.
Based on this observation, CAPTA focuses on model preferences rather than capabilities.

Each CAPTA probe consists of a simple question and a set of multiple reasonable options.
During verification, different LLMs may select different options due to their distinct preferences, allowing the remote service to identify the backend LLM.
However, an LLM may not always select the same option across repeated trials, as its responses can vary.
Therefore, CAPTA captures model preferences as probability distributions rather than fixed choices.

Moreover, model preferences may be influenced by the order in which options are presented.
For example, some models may favor options appearing earlier in the list.
CAPTA does not attempt to separate such positional effects from other preference factors, but treats the resulting selection behavior as part of the model's preference fingerprint.
This is reasonable because CAPTA uses identical probes, including option orders, during preparation and runtime identification, ensuring that consistent positional biases are also captured in the reference fingerprints.

% \sisi{consider discussing whether LLMs always answer the same  to the same question.}

Since each probe consists of a question and an option set, probe construction involves two challenges:

\begin{table}[tp]
\centering
\caption{Changing the question reveals differences in model preferences.}
\label{tab:preference-wording}
\begingroup
% Original monochrome vector pictograms for the paper's table headers.
% Illustrative only: the experimental prompts contain animal names, not images.
% All six icons share a 32 x 32 canvas, rounded strokes and a fixed display size.
\newcommand{\captaanimalicon}[1]{%
  \resizebox{!}{3.6mm}{%
  \begin{tikzpicture}[x=1pt,y=1pt,line width=1.2pt,
    line cap=round,line join=round,draw=black,fill=white]
  \path[use as bounding box] (0,0) rectangle (32,32);
  \csname captaanimaldraw#1\endcsname
  \end{tikzpicture}}%
}

\newcommand{\captaanimaldrawzebra}{%
  \draw[fill=white] (9,23) .. controls (4,31) and (10,32) .. (12,25)
    (20,25) .. controls (23,32) and (28,31) .. (23,22);
  \draw[fill=white] (10,25) .. controls (12,29) and (21,29) .. (23,24)
    -- (25,10) .. controls (25,4) and (8,4) .. (8,10) -- cycle;
  \draw (10,23)--(15,20)--(11,19) (22,23)--(18,20)--(23,18)
    (10,17)--(14,15) (23,16)--(19,14);
  \draw[fill=white] (8,10) .. controls (11,14) and (22,14) .. (25,10);
  \fill[black] (12,17.5) circle (1) (21,17.5) circle (1)
    (12,9) circle (.8) (21,9) circle (.8);
}
\newcommand{\captaanimaldrawgiraffe}{%
  \draw (12,26)--(11,30) (20,26)--(21,30);
  \fill[black] (11,30) circle (1.4) (21,30) circle (1.4);
  \draw[fill=white] (10,25) .. controls (2,29) and (3,21) .. (10,21)
    (22,25) .. controls (30,29) and (29,21) .. (22,21);
  \draw[fill=white] (12,14)--(11,2)--(21,2)--(20,14);
  \draw[fill=white] (10,24) .. controls (12,27) and (20,27) .. (22,24)
    --(23,16) .. controls (22,11) and (10,11) .. (9,16)--cycle;
  \draw (10,16) .. controls (14,18) and (19,18) .. (22,16);
  \fill[black] (12.5,21) circle (1) (19.5,21) circle (1)
    (14,9) circle (1.5) (18.5,5) circle (1.4);
}
\newcommand{\captaanimaldrawlion}{%
  \draw[fill=white] (16,31)--(20,28)--(25,29)--(26,24)--(30,21)
    --(28,16)--(30,11)--(25,9)--(23,4)--(18,5)--(16,1)
    --(12,5)--(7,4)--(6,9)--(2,12)--(4,17)--(2,22)
    --(7,25)--(8,29)--(13,28)--cycle;
  \draw[fill=white] (8,22) circle (3) (24,22) circle (3);
  \draw[fill=white] (16,17) ellipse (9 and 10);
  \fill[black] (12,20) circle (1) (20,20) circle (1);
  \draw[fill=black] (13.5,15)--(18.5,15)--(16,12.5)--cycle;
  \draw (16,12.5)--(16,10) (12,11) .. controls (13,9) and (15,9) .. (16,10)
    .. controls (17,9) and (19,9) .. (20,11);
}
\newcommand{\captaanimaldrawbuffalo}{%
  \draw[fill=white] (12,23) .. controls (5,19) and (1,24) .. (3,30)
    .. controls (5,25) and (8,27) .. (13,27)
    --(19,27) .. controls (24,27) and (27,25) .. (29,30)
    .. controls (31,24) and (27,19) .. (20,23);
  \draw[fill=white] (10,23)--(4,22)--(7,18)--(10,19)
    (22,23)--(28,22)--(25,18)--(22,19);
  \draw[fill=white] (10,24) .. controls (13,26) and (19,26) .. (22,24)
    --(23,12) .. controls (23,3) and (9,3) .. (9,12)--cycle;
  \draw (10,11) .. controls (13,14) and (20,14) .. (22,11);
  \fill[black] (12,18) circle (1) (20,18) circle (1)
    (13,9) circle (.9) (19,9) circle (.9);
}
\newcommand{\captaanimaldrawrhinoceros}{%
  \draw[fill=white] (9,23) .. controls (5,31) and (14,31) .. (13,24);
  \draw[fill=white] (5,8)--(5,19) .. controls (8,27) and (18,25) .. (20,19)
    --(23,15)--(28,12)--(30,7) .. controls (26,3) and (13,4) .. (10,7)--cycle;
  \draw[fill=white] (21,14) .. controls (25,19) and (26,22) .. (26,26)
    .. controls (30,19) and (28,15) .. (27,12)--cycle;
  \draw[fill=white] (18,17)--(20,22)--(22,15);
  \fill[black] (15,18) circle (1) (27,8) circle (.8);
  \draw (20,6)--(28,6);
}
\newcommand{\captaanimaldrawelephant}{%
  \draw[fill=white] (11,25) .. controls (-1,32) and (0,7) .. (9,8)
    --(13,16) (21,25) .. controls (33,32) and (32,7) .. (23,8)--(19,16);
  \draw[fill=white] (10,25) .. controls (12,30) and (20,30) .. (22,25)
    --(22,15)--(19,11)--(19,5)
    .. controls (19,0) and (11,0) .. (10,5)
    --(13,6) .. controls (13,3) and (16,3) .. (16,6)
    --(15,13)--(10,16)--cycle;
  \draw (10,14)--(8,8)--(13,12) (22,14)--(24,8)--(19,12);
  \fill[black] (12,20) circle (1) (20,20) circle (1);
}
\footnotesize
\setlength{\tabcolsep}{2pt}
\renewcommand{\arraystretch}{1.15}
\begin{tabular*}{\linewidth}
  {@{\extracolsep{\fill}}l
   *{5}{>{\centering\arraybackslash}p{23pt}}
   >{\centering\arraybackslash}p{30pt}@{}}
\toprule
      & \makecell[c]{\captaanimalicon{zebra}\\zebra}
      & \makecell[c]{\captaanimalicon{giraffe}\\giraffe}
      & \makecell[c]{\captaanimalicon{lion}\\lion}
      & \makecell[c]{\captaanimalicon{buffalo}\\buffalo}
      & \makecell[c]{\captaanimalicon{rhinoceros}\\rhino}
      & \makecell[c]{\captaanimalicon{elephant}\\elephant} \\
\midrule

\multicolumn{7}{@{}l}{%
  \textbf{Question (a): \emph{Which animal is the most iconic symbol of Africa?}}} \\
\addlinespace[2pt]

\evalmodel{openai}{GPT-5.6-Luna}
  & 0 & 0 & \cellcolor{black!8}\textbf{20} & 0 & 0 & 0 \\
\evalmodel{claude}{Claude-Sonnet-5}
  & 0 & 0 & \cellcolor{black!8}\textbf{20} & 0 & 0 & 0 \\
\evalmodel{gemini}{Gemini-3.7-Flash}
  & 0 & 0 & \cellcolor{black!8}\textbf{20} & 0 & 0 & 0 \\
\evalmodel{kimi}{Kimi-K2.6}
  & 0 & 0 & \cellcolor{black!8}\textbf{20} & 0 & 0 & 0 \\

\midrule

\multicolumn{7}{@{}l}{%
  \textbf{Question (b): \emph{Choose one African animal.}}} \\
\addlinespace[2pt]

\evalmodel{openai}{GPT-5.6-Luna}
  & 0 & \cellcolor{black!8}\textbf{20} & 0 & 0 & 0 & 0 \\
\evalmodel{claude}{Claude-Sonnet-5}
  & 0 & \cellcolor{black!8}2 & 0 & 0 & 0 & \cellcolor{black!8}\textbf{18} \\
\evalmodel{gemini}{Gemini-3.7-Flash}
  & \cellcolor{black!8}\textbf{17} & 0 & \cellcolor{black!8}3 & 0 & 0 & 0 \\
\evalmodel{kimi}{Kimi-K2.6}
  & 0 & 0 & \cellcolor{black!8}\textbf{20} & 0 & 0 & 0 \\

\bottomrule
\end{tabular*}
\endgroup

\par\smallskip
\parbox{\linewidth}{\footnotesize
Each question is evaluated with 20 responses per model.}
\end{table}

% Source: results/24-game-piece-preference/formal-04-seven-model-search/
% report.json and collection.json: r14_confirm_baseline / r15_confirm_endpoint.
% One deletion step, two measured stages; 20 responses per model per stage.

\begin{table}[tp]
\centering
\caption{Removing a shared preferred option reveals differences in model preferences.}
\label{tab:option-deletion}

\footnotesize
\setlength{\tabcolsep}{1.5pt}
\renewcommand{\arraystretch}{1.15}

\parbox{\linewidth}{
\textbf{Question:}
\emph{You are designing a board game.
Select one classic shape for its player tokens.}
}
\par\smallskip

\begin{tabular*}{\linewidth}
{@{\extracolsep{\fill}}lccccccc@{}}
\toprule
& shoe & cube & car & meeple & knight & thimble & cone \\
\midrule

\multicolumn{8}{@{}l}{
    \textbf{(a) \textit{Original option set}}
} \\
\addlinespace[2pt]

\evalmodel{openai}{GPT-5.6-Luna}
    & 0 & 0 & 0 & \cellcolor{black!8}\textbf{20} & 0 & 0 & 0 \\

\evalmodel{claude}{Claude-Sonnet-5}
    & 0 & 0 & 0 & \cellcolor{black!8}\textbf{20} & 0 & 0 & 0 \\

\evalmodel{gemini}{Gemini-3.7-Flash}
    & 0 & 0 & 0 & \cellcolor{black!8}\textbf{20} & 0 & 0 & 0 \\

\evalmodel{moonshot}{Kimi-K2.6}
    & 0 & 0 & 0 & \cellcolor{black!8}\textbf{20} & 0 & 0 & 0 \\

\midrule

\multicolumn{8}{@{}l}{
     \textbf{(b) \textit{After removing meeple}}
} \\
\addlinespace[2pt]

\evalmodel{openai}{GPT-5.6-Luna}
    & 0 & \cellcolor{black!8}1 & 0 & -- & \cellcolor{black!8}\textbf{19} & 0 & 0 \\

\evalmodel{claude}{Claude-Sonnet-5}
    & 0 & 0 & 0 & -- & 0 & \cellcolor{black!8}\textbf{20} & 0 \\

\evalmodel{gemini}{Gemini-3.7-Flash}
    & 0 & \cellcolor{black!8}1 & 0 & -- & 0 & 0 & \cellcolor{black!8}\textbf{19} \\

\evalmodel{moonshot}{Kimi-K2.6}
    & 0 & 0 & \cellcolor{black!8}\textbf{20} & -- & 0 & 0 & 0 \\

\bottomrule
\end{tabular*}

\par\smallskip
\parbox{\linewidth}{\footnotesize
Each option set is tested 20 times per model; ``--'' denotes removal.
}

\end{table}

\bheading{Question generation and selection challenge.}
The question itself can significantly affect model preferences.
As shown in Table~\ref{tab:preference-wording}, the same option set can lead to different preferences under different questions.
In Question (a), all four models consistently select \textit{lion}, making them indistinguishable.
Meanwhile, Question (b) leads to different choices across models, making them easier to distinguish.
It is important to find questions that elicit different model preferences.
However, the space of possible questions is vast, making it difficult to identify promising ones efficiently.
This leads to our first challenge:

\challenge{
\label{ch:question-selection}
How can we efficiently identify promising questions that elicit different preferences across LLMs?
}

\bheading{Option-set optimization challenge.}
For the same question, different option sets may lead to different model preferences.
As shown in Table~\ref{tab:option-deletion}, all four models select \textit{meeple} with the original option set across 20 trials.
After \textit{meeple} is removed, the models begin to select different options.
The number of possible option sets can be very large. For example, an initial set of 30 options has approximately $1.07\times10^9$ possible subsets.
This leads to our second challenge:

\challenge{
\label{ch:option-optimization}
How can we efficiently find a high-quality option set from a large search space?
}

\subsection{CAPTA Solutions and Workflow}

As shown in Figure~\ref{fig:design-overview}, CAPTA consists of two phases.
In the preparation phase, CAPTA constructs probes and builds a reference fingerprint database through three stages.
In the runtime phase, CAPTA uses the probes and fingerprints to identify the backend LLM of a target agent.

\bheading{Stage 1: Question generation and selection.}
CAPTA first generates candidate questions and uses multiple LLMs to generate valid and invalid options for each question.
It then aggregates similar options and ranks questions based on the diversity of their valid options, selecting the top-ranked questions for the next stage to address Challenge 1.
The intuition is that questions for which different LLMs generate more diverse options may have greater potential to distinguish their preferences.

\bheading{Stage 2: Option-set optimization.}
For each selected question, CAPTA constructs an initial option set using the options generated in Stage 1.
It then uses an LLM as an optimizer to iteratively remove options and find a more discriminative set, addressing Challenge 2.
In each round, CAPTA queries multiple LLMs and evaluates the current option set based on their preference distributions.
The evaluation results are passed to the optimizer, which decides whether to accept the changes or roll back.
The resulting optimized probes are then used for fingerprint construction.

\bheading{Stage 3: Reference fingerprint construction.}
CAPTA repeatedly queries candidate LLMs with the optimized probes and records their selected options.
For each model and probe, it estimates a probability distribution based on the observed responses.
These distributions capture each model's preference patterns.
The remote service stores them as reference fingerprints for later identification.

\bheading{Stage 4: Likelihood-based identification.}
During runtime, the remote service presents a set of CAPTA probes to the target agent as CAPTCHA challenges and collects its responses.
CAPTA then matches the collected responses against the reference fingerprints to identify the backend LLM.
If multiple agent--service interactions are available, CAPTA can collect additional responses to further improve identification accuracy.

\begin{table}[t]
\centering
\caption{CAPTA's handling of different scenarios.}
\label{tab:model-maintenance}
\begingroup
\footnotesize
\setlength{\tabcolsep}{1pt}
\setlength{\extrarowheight}{0pt}
\renewcommand{\arraystretch}{1.25}
\renewcommand{\tabularxcolumn}[1]{m{#1}}

\begin{tabularx}{\linewidth}{
>{\hsize=0.92\hsize\linewidth=\hsize\raggedright\arraybackslash}X
>{\hsize=1.90\hsize\linewidth=\hsize\raggedright\arraybackslash}X
>{\hsize=0.48\hsize\linewidth=\hsize\centering\arraybackslash}X
>{\hsize=0.7\hsize\linewidth=\hsize\centering\arraybackslash}X}
\toprule
\textbf{Scenario}
& \textbf{CAPTA action}
& \makecell{\textbf{New}\\\textbf{probes}}
& \makecell{\textbf{New}\\\textbf{fingerprints}} \\
\midrule

% \rowcolor{gray!10}[0pt][0pt]
% \multicolumn{4}{@{}l@{}}{\textbf{Initial setup}} \\
% \addlinespace[2pt]

% Candidate models
% & Construct and optimize probes; collect reference fingerprints
% & Yes
% & Yes \\

% \midrule

\rowcolor{gray!10}[0pt][0pt]
\multicolumn{4}{@{}l@{}}{\textbf{Candidate models with existing fingerprints}} \\
\addlinespace[2pt]

Different tasks
\par\smallskip
\hrule width\linewidth height\cmidrulewidth\relax
\smallskip
Different agents
& Reuse probes and fingerprints; collect additional runtime responses if needed
& No
& No \\

\midrule
\rowcolor{gray!10}[0pt][0pt]
\multicolumn{4}{@{}l@{}}{\textbf{New models without existing fingerprints}} \\
\addlinespace[2pt]

Newly released
\par\smallskip
\hrule width\linewidth height\cmidrulewidth\relax
\smallskip
Model updates
& Treat all as new candidates; reuse probes and collect fingerprints; add probes only if needed.
& If needed
& Yes \\

\cmidrule{1-4}

Unknown models
& Apply open-set identification to detect and reject unknown models
& No
& No \\

\bottomrule
\end{tabularx}
\endgroup
\end{table}

Table~\ref{tab:model-maintenance} summarizes how CAPTA handles different scenarios.
CAPTA first constructs and optimizes probes based on an initial set of candidate models and collects reference fingerprints.
Since a remote service typically knows its own service and task, it can provide the corresponding task context when collecting fingerprints.
Nevertheless, our experiments show that the same fingerprints can be reused across different tasks and agents.
The service does not need to access the agent's internal information, such as system prompts, tool interfaces, or execution contexts.
Although model preferences may shift with context changes, collecting additional runtime responses can mitigate their impact on identification accuracy.

For newly released models and updated model versions that are to be added to the candidate set, CAPTA treats each as a new candidate, reuses existing probes, and collects new reference fingerprints.
Our experiments show that CAPTA can distinguish models both across and within model families and remains effective under different runtime temperatures.
For difficult models, CAPTA can construct additional probes or collect more runtime responses to improve identification accuracy.
Finally, CAPTA applies open-set identification to detect and reject unknown models.

\section{CAPTA Design}
\label{sec:design}

Let $\mathcal{M}=\{m_1,\ldots,m_K\}$ denote the set of candidate LLMs.
A CAPTA probe $p=(q,O)$ consists of a question $q$ and an option set $O$.
CAPTA aims to construct probes that reveal differences in preferences among candidate LLMs.
This section describes the four stages of CAPTA.

\subsection{Question Generation and Selection}
\label{sec:question-design}

Stage 1 addresses Challenge 1 by ranking candidate questions based on the diversity of their generated options.
We then select the top-ranked questions for the next stage.

\bheading{Question design and generation.}
A suitable question should provide multiple reasonable and distinct answers, while avoiding wording that favors a particular choice. 
First, a richer choice space gives models more opportunities to express different preferences. For example, a rainbow question offers seven reasonable answers, compared with only three for a traffic-light question. 
Second, questions should use simple and neutral wording rather than suggest preferred answers through terms such as \textit{best} or \textit{most}. 
As shown in Table~\ref{tab:preference-wording}, such wording leads all four models to select \textit{lion}, while neutral wording elicits more diverse choices.

Based on these principles, CAPTA uses LLMs to generate candidate questions.
This process produces $N_{\mathrm{cand}}$ candidate questions, $\mathcal{Q}=\{q_1,q_2,\ldots,q_{N_{\mathrm{cand}}}\}$.

\bheading{Question selection via option diversity.}
Model discrimination depends on both the question and its option set. 
Recall that different option sets can lead to different preference patterns even for the same question (Table~\ref{tab:option-deletion}).
Therefore, directly evaluating candidate questions based on model preferences would require first determining suitable option sets, which is addressed in Stage 2. 
Instead, CAPTA uses option diversity to rank candidate questions.
The intuition is that if different LLMs generate similar options for a question, they may also exhibit similar preferences when answering it.
We therefore use option overlap to estimate a question's potential to distinguish models.

Specifically, for each candidate question $q$, we use a set of LLMs $\mathcal{G} \subseteq \mathcal{M}$ to independently generate valid and invalid options.
We aggregate the generated options separately, merging obvious surface-form variants. 
For example, \textit{lion}, \textit{lions}, and \textit{a lion} are treated as the same option. 
This gives us a valid option set $V_q$ and an invalid option set $I_q$.

We also record which LLMs generated each option.
When calculating the diversity score, we consider only valid options $V_q$, as LLMs are more likely to select valid options than invalid ones.
For each valid option $o\in V_q$, its cross-model generation frequency is defined as
\[
f_q(o)=
\left|
\{g\in\mathcal{G}\mid g\text{ generated }o\}
\right|,
\]
where $|\mathcal{G}|$ is the number of LLMs used for option generation.
Each LLM is counted at most once for the same option. Thus,
$ 1\leq f_q(o)\leq |\mathcal{G}|. $

Based on these frequencies, we define the diversity score of a question $q$ as
\[
D(q)=
1-
\frac{1}{|V_q|}
\sum_{o\in V_q}
\frac{f_q(o)}{|\mathcal{G}|}.
\]
The score satisfies $D(q)\in[0,1)$.
A higher score means less overlap among the valid options generated by different LLMs and thus greater option diversity.

CAPTA ranks candidate questions in descending order of $D(q)$ and selects the top-ranked questions for Stage 2.

\subsection{Option-Set Optimization}
\label{sec:option-optimization}

In Stage 2, CAPTA constructs initial option sets and uses an LLM-guided search to iteratively remove options, addressing Challenge 2.
The optimized probes are then used to construct reference fingerprints.

\bheading{Initial probe construction.}
For each selected question $q$, CAPTA ranks valid options by their generation frequencies $f_q(o)$, with less frequently generated options ranked higher. It then selects a number of valid options from the ranked list while preserving their relative order.
As described before, we also treat positional preferences, such as favoring earlier options, as part of a model's preference fingerprint.
This ordering places less frequently generated options earlier to reduce the potential dominance of commonly preferred options.
CAPTA also randomly samples a small number of invalid options and inserts them at random positions in the sequence, forming the initial option set $O$.
Note that CAPTA focuses on model preferences rather than answer correctness. 
In other words, CAPTA accepts and records all model responses, including selections of invalid options, for fingerprint construction.

However, the initial option set may not provide sufficient model discrimination. 
For an initial option set containing $L$ options, CAPTA keeps the relative option order fixed and considers all subsets containing at least two options. This results in $\sum_{k=2}^{L}\binom{L}{k}=2^L-L-1$ possible subsets.
For example, with 30 options, the search space contains approximately $1.07\times10^9$ subsets. Moreover, each subset creates a different choice context and requires new queries to estimate preference distributions. 
Therefore, exhaustive search is infeasible.

\bheading{LLM-guided optimization.}
For a given probe $p=(q,O)$, CAPTA utilizes a heuristic method to optimize its option set.
To reduce sampling costs, CAPTA uses a representative subset $\mathcal{M}'\subset\mathcal{M}$ of candidate LLMs during optimization.
For each probe, CAPTA repeatedly queries the representative models to estimate their preference distributions over the options.
We aim to find option sets that provide discriminative and reliable evidence from responses.

CAPTA uses a separate LLM as an optimizer to guide the search.
The optimizer makes two main decisions: (1) which options to remove and (2) whether to accept or roll back the resulting changes.
To focus on model preference patterns rather than option semantics, CAPTA assigns anonymous IDs to models and options and excludes the question and option texts from the optimizer's input.

The optimizer also receives evaluation metrics, a preference probability table, and search history.
Appendix~\ref{app:optimizer-inputs} provides a detailed description of these inputs. We briefly introduce them below.

Specifically, CAPTA uses three metrics:
\begin{packeditemize}
    \item $MI_{\mathrm{ratio}}$ measures how much information the responses provide about model identity~\cite{DBLP:conf/acl/SorensenRRSRDKF22, cover2006elements}, normalized by the maximum entropy of model identity.

    \item $A_p$ measures the expected single-probe identification accuracy under a uniform prior over candidate models.

    \item $R_2$ measures the average squared dominant-choice probability across models, favoring option sets where each model has a stronger preference for one option.
\end{packeditemize}
These metrics help the optimizer evaluate whether an optimization improves or degrades the option set and decide whether to accept the changes or roll them back.

The probability table records the probability of each model selecting each option, helping the optimizer identify options for removal.
The optimizer explores two directions: removing options frequently selected by multiple models to examine whether their preferences diverge, and removing options rarely selected by a particular model to examine whether its preference for another option becomes stronger.

After each proposed removal, CAPTA constructs a new probe and queries the representative models for reevaluation.
The optimizer accepts the modification only if at least two of the three metrics improve; otherwise, it rolls back.
The search history records the removed options and acceptance decisions in each round, helping the optimizer track previous results and restore earlier states when needed.

Throughout the search, CAPTA separately tracks the best-performing option set $O^\star$ found so far. When the search budget is exhausted or a stopping condition is reached, CAPTA returns the optimized probe $p=(q,O^\star)$.

\subsection{Fingerprint Construction and Identification}
\label{sec:fingerprint-identification}

In Stage 3, CAPTA queries each candidate LLM with every optimized probe and uses the option-selection frequencies to construct reference fingerprints.
In Stage 4, CAPTA collects responses from the target agent and matches them against the fingerprints to identify its backend LLM.

Recall that $\mathcal{M}=\{m_1,m_2,\ldots,m_K\}$ denotes the set of $K$ candidate LLMs.
Let $\mathcal{P}=\{p_1,p_2,\ldots,p_N\}$ denote the set of $N$ optimized probes.
Each probe $p_i=(q_i,O_i^\star)$ consists of a question $q_i$ and an optimized option set $O_i^\star$.
We formalize the processes in Stages 3 and 4 below.

\bheading{Preference distribution.}
For a model $m$, let $X_i$ denote the option selected by the model when answering probe $p_i$.
A model may select different options when answering the same probe repeatedly.
We therefore use a probability distribution to describe how likely the model is to select each option:
\[
P_{m,i}(o)=\Pr(X_i=o\mid m,p_i),
\qquad o\in O_i^\star.
\]
For example, if a model selects option $o_{1}$ in 80\% of repeated
tests and option $o_{2}$ in the remaining 20\%, their probabilities
are 0.8 and 0.2, respectively.
%CAPTA uses these differences in option preferences to distinguish models.

\bheading{Reference fingerprint.}
During preparation, CAPTA tests each probe $T_{\mathrm{prep}}$ times on each candidate model.
Let $c_{m,i,o}$ denote the number of times model $m$ selects option $o$
when answering probe $p_i$.
We estimate the probability of selecting each option based on the observed selection counts:
\[
\hat{P}_{m,i}(o)
=
\frac{c_{m,i,o}+\alpha}{T_{\mathrm{prep}}+\alpha J_i},
\]
where $\alpha=1$ is the Laplace smoothing parameter and $J_i=|O_i^\star|$
is the number of options for probe $p_i$.
Laplace smoothing adds $\alpha$ to each option's count, 
ensuring that every option has a nonzero estimated probability, 
even if it was never selected during preparation.

For each candidate model $m$, its reference fingerprint consists of the preference distributions across all probes:
$
F_m=\{\hat{P}_{m,1},\hat{P}_{m,2},\ldots,\hat{P}_{m,N}\}
$.
All reference fingerprints form the fingerprint database:
$
\mathcal{F}=\{F_m\mid m\in\mathcal{M}\}.
$

\bheading{Runtime identification.}
During runtime, CAPTA typically may collect multiple responses.
Let $T_i$ denote the number of responses collected for probe $p_i$,
and let $x_{i,t}$ denote the option selected in the $t$-th response,
where $x_{i,t}\in O_i^\star$.

We denote the responses collected for probe $p_i$ by
$\mathbf{x}_i=(x_{i,1},\ldots,x_{i,T_i})$,
and all runtime observations across the $N$ probes by
$\mathbf{x}=(\mathbf{x}_1,\ldots,\mathbf{x}_N)$.
CAPTA identifies the backend LLM by matching the observed responses against the fingerprint database:
$
\hat{m}=f(\mathbf{x},\mathcal{F}),
$
where $f$ denotes the identification function.

\bheading{Likelihood-based identification.}
We use likelihood-based matching~\cite{DBLP:conf/icml/RennieSTK03} as the identification function $f$.
Assuming that responses are conditionally independent given the backend model and probes, the likelihood of observing $\mathbf{x}$ under model $m$ is estimated as
\[
\hat{P}(\mathbf{x}\mid m)
=
\prod_{i=1}^{N}
\prod_{t=1}^{T_i}
\hat{P}_{m,i}(x_{i,t}),
\]
where $N$ is the number of probes and $T_i$ is the number of responses collected for probe $p_i$.

To avoid multiplying small probabilities, we sum the log-likelihoods across all observed responses and identify the model with the highest total score:
\[
\hat{m}
=
\arg\max_{m\in\mathcal{M}}
\sum_{i=1}^{N}
\sum_{t=1}^{T_i}
\log\hat{P}_{m,i}(x_{i,t}).
\]

\section{Evaluation}
\label{sec:eval}

We conduct extensive experiments to evaluate CAPTA's effectiveness and practical feasibility, as shown in Table~\ref{tab:eval-overview}.

\subsection{Experimental Setup}
\label{sec:eval-setup}

\bheading{Models.}
We refer to existing LLM benchmarks~\cite{lmarena, artificialanalysis, openrouter} to select popular models.
Our evaluation includes 25 LLMs from 10 model families as listed in Table~\ref{tab:eval-model-accuracy}.
These range from high-performance models to low-cost models, all of which may serve as backend LLMs for agents.
We select seven representative models, marked with $^\ast$ in the table, for probe construction in Stages 1 and 2.

\bheading{Agents.}
We evaluate CAPTA on four popular agents: OpenClaw~\cite{openclaw_runtime}, Hermes~\cite{hermes_agent}, Cline~\cite{cline}, and DeepSeek Harness~\cite{deepseek_harness}.
OpenClaw has over 12.6M monthly downloads,
Cline has over 5.9M installations,
and both Hermes and DeepSeek Harness have over 200K GitHub stars.
% For each agent, we configure the seven representative LLMs as its backend and expose the remote weather service through an agent-specific skill.
% We use the same weather-query task across all agents, with each agent--LLM combination running in an independent session.

\bheading{CAPTA settings.}
In Stage 1, we use GPT-5.6-Terra and Claude-Sonnet-5 to generate candidate questions and GPT-5.6-Terra for deduplication, resulting in 275 candidate questions.
We use the seven representative models to generate options and rank the questions by their diversity scores.

In Stage 2, we select the top 30 questions and use seven representative models for sampling, with GPT-5.6-Terra as the optimizer. 
For each question, we construct an initial option set with up to 30 valid options, and randomly add invalid options equal to 10\% of the selected valid options. 
We set a maximum of 12 optimization rounds and stop early if no improvement occurs for six consecutive rounds.
%When more are available, we sample proportionally from the top 25\%, middle 50\%, and bottom 25\% of the generation-frequency ranking

\bheading{Stage 3: Fingerprint collection.}
We collect reference fingerprints in three settings, with separate independent queries at each model's default temperature.
For the identification evaluation in Section~\ref{sec:eval:identification}, we collect JSON responses for 30 probes across all 25 LLMs.
For the agent evaluation in Section~\ref{sec:eval:agent}, we use seven representative LLMs and the top-10 probes, submitting selections through \texttt{web\_fetch} with the task description provided.
For the cross-task evaluation in Section~\ref{sec:eval:cross-task}, we collect task-independent fingerprints using the same seven models and top-10 probes in a neutral resource-query setting.

\bheading{Stage 4: Identification.}
For the main evaluation, we use repeated random holdout~\cite{DBLP:conf/uss/Li0FK21, DBLP:conf/uss/ZuffereyHTH23}.
In each trial, one response is randomly selected for runtime identification, while the remaining 50 are used to construct the reference fingerprint.
This design provides sufficient samples for reliable fingerprint estimation while avoiding the bias of a fixed train--test split.
For each probe budget, we repeat this process 1,000 times and report the mean identification accuracy.

\begin{table}[t]
\centering
\caption{Overview of CAPTA's evaluation.}
\label{tab:eval-overview}
\begingroup
\footnotesize
\setlength{\tabcolsep}{1pt}
\setlength{\extrarowheight}{0pt}
\renewcommand{\arraystretch}{1.25}
\renewcommand{\tabularxcolumn}[1]{m{#1}}

\begin{tabularx}{\linewidth}{
>{\hsize=1.30\hsize\linewidth=\hsize\raggedright\arraybackslash}X
>{\hsize=0.70\hsize\linewidth=\hsize\raggedright\arraybackslash}X}
\toprule
\textbf{Evaluation}
& \textbf{Main Evidence} \\
\midrule

\textbf{Identification capability}
(\S\ref{sec:eval:identification})
\par
Can CAPTA identify backend LLMs?
& 25 LLMs; 30 probes; same-family models \\

\addlinespace[2pt]

\rowcolor{gray!8}
\textbf{Probe construction}
(\S\ref{sec:eval:effectiveness})
\par
Are our probe construction methods effective?
& Question ranking; baseline vs. optimized \\

\addlinespace[2pt]

\textbf{Agent evaluation}
(\S\ref{sec:eval:agent})
\par
How does CAPTA perform in practice?
& 4 agents; 7 LLMs; total 28 combinations \\

\addlinespace[2pt]

\rowcolor{gray!8}
\textbf{Fingerprint reuse and context effects}
(\S\ref{sec:eval:cross-task})
\par
Can CAPTA reuse fingerprints across tasks?
& Independent fingerprints tested on different tasks \\

\addlinespace[2pt]

\textbf{Open-set identification}
(\S\ref{sec:eval:open-set})
\par
Can CAPTA detect unknown models?
& Unknown/known models \\

\addlinespace[2pt]

\rowcolor{gray!8}
\textbf{Practical considerations}
(\S\ref{sec:eval:practical})
\par
What practical factors affect deployment?
& Submission; sampling; temperature; token cost \\

\bottomrule
\end{tabularx}
\endgroup
\end{table}

\subsection{Backend LLM Identification}
\label{sec:eval:identification}

\bheading{Reliability of self-reported identity.}
Before evaluating CAPTA, we first consider a more direct approach:
asking a model to report its own identity.
We query each of the 25 models 10 times, obtaining 250 responses in total.
Only 20 responses (\accp{8.00}) correctly report the exact model identity.
Moreover, some models consistently report incorrect identities. 
For example, DeepSeek-V4-Flash reports itself as \textit{gpt-4o-mini} in all 10 trials.
Appendix~\ref{app:eval-self-report} provides more details.

\finding{
Although some models correctly identify themselves, others consistently report incorrect identities, making self-reports unreliable for model identification.
}

\begin{figure}[tp]
\centering
\includegraphics[width=0.95\linewidth]{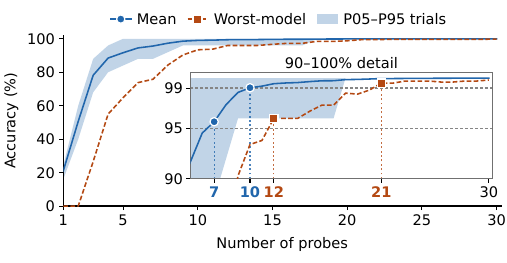}
\caption{
    Identification accuracy across 25 models. 
    CAPTA reaches 99\% mean accuracy with 10 probes and 99\% worst-model accuracy with 21 probes.
}
\label{fig:eval-accuracy}
\end{figure}

\bheading{Accuracy under different probe budgets.}
We rank probes by their diversity scores and progressively combine the top-ranked probes to evaluate identification accuracy.
As shown in Figure~\ref{fig:eval-accuracy}, CAPTA's mean accuracy increases rapidly with the number of probes, reaching 95\% with 7 probes and 99\% with 10 probes.
The worst-model accuracy reaches 95\% with 12 probes and 99\% with 21 probes.

Table~\ref{tab:eval-model-accuracy} further shows the identification accuracy across individual models.
With only 5 probes, CAPTA achieves a mean accuracy of \accp{91.65}, but some models, such as DeepSeek-V4-Flash, Grok-4.6, and GLM-5.3, achieve only \accp{64.75}, \accp{67.20}, and \accp{68.00}, respectively.
Increasing the probe budget enables more models to achieve high accuracy.
For example, with 15 probes, all 25 models achieve at least 95\% accuracy, while 22 models reach 99\%.

In practice, a single Google reCAPTCHA verification may present users with 5--10 image-selection challenges.
CAPTA can adopt a similar mechanism to present multiple probes in a single interaction, reducing the number of interactions needed for identification.

\bheading{Distinguishing models within the same family.}
Table~\ref{tab:eval-model-accuracy} shows that CAPTA can distinguish not only models from different families but also models within the same family.
For example, with 10 probes, most models from the OpenAI, Claude, Gemini, and Qwen families achieve over \accp{99.80} accuracy.
These results indicate that models within the same family can exhibit distinct preference patterns.

\bheading{Identifying difficult models.}
However, some models require more probes to achieve high accuracy.
For example, Claude-Opus-5 reaches 99\% accuracy with only one probe, whereas DeepSeek-V4-Pro, Grok-4.5, and Grok-4.6 require 19, 18, and 21 probes, respectively.
Grok-4.5 and Grok-4.6 achieve \accp{93.4} and \accp{95.0} accuracy with 10 probes, respectively.
Specifically, \accp{6.00} of Grok-4.5 responses are misidentified as Grok-4.6, while \accp{4.60} of Grok-4.6 responses are misidentified as Grok-4.5.
With 21 probes, these error rates decrease to \accp{0.20} and \accp{0.50}, respectively.

\finding{
CAPTA achieves 99\% mean accuracy with 10 probes and 99\% worst-model accuracy with 21 probes.
It distinguishes models within the same family, while more probes help identify difficult models.
}

\bheading{Generalization to new models.}
Although Stages 1 and 2 use only seven representative models for probe construction, we apply the optimized probes to 18 additional models in Stage 3 and construct their reference fingerprints, expanding the candidate set to 25 models.
CAPTA achieves similar identification accuracy on these two groups.
As shown in Table~\ref{tab:eval-model-accuracy}, the mean accuracy differs by at most 0.9 percentage points between the two groups across different probe budgets.
With 10 probes, the seven representative models and 18 additional models achieve \accp{99.70} and \accp{98.77} mean accuracy, respectively, and both approach 100\% with 30 probes.

\finding{
CAPTA generalizes well to new models without repeating probe construction; new models only require collecting their reference fingerprints.
}

\begin{table}[tp]
\centering
\caption{Identification accuracy (\%) across 25 models using the top-$N$ probes ranked by diversity score.}
\label{tab:eval-model-accuracy}
\footnotesize
\setlength{\tabcolsep}{1.5pt}
\renewcommand{\arraystretch}{1.15}
\begin{tabular*}{\linewidth}{@{\extracolsep{\fill}}lrrrrrr@{}}
\toprule
 & \multicolumn{6}{c}{\textbf{Number of probes}} \\
\cmidrule(lr){2-7}
\textbf{Model} & \textbf{5} & \textbf{10} & \textbf{15} & \textbf{20} & \textbf{25} & \textbf{30} \\
\midrule
\evalmodel{openai}{GPT-5.6-Luna}$^{*}$ & \acccell{100.00} & \acccell{100.00} & \acccell{100.00} & \acccell{100.00} & \acccell{100.00} & \acccell{100.00} \\
\evalmodel{openai}{GPT-5.6-Terra} & \acccell{89.90} & \acccell{99.80} & \acccell{100.00} & \acccell{100.00} & \acccell{100.00} & \acccell{100.00} \\
\evalmodel{openai}{GPT-5.6-Sol} & \acccell{99.90} & \acccell{100.00} & \acccell{100.00} & \acccell{100.00} & \acccell{100.00} & \acccell{100.00} \\
\addlinespace[2pt]
\evalmodel{claude}{Claude-Sonnet-5}$^{*}$ & \acccell{100.00} & \acccell{100.00} & \acccell{100.00} & \acccell{100.00} & \acccell{100.00} & \acccell{100.00} \\
\evalmodel{claude}{Claude-Opus-5} & \acccell{100.00} & \acccell{100.00} & \acccell{100.00} & \acccell{100.00} & \acccell{100.00} & \acccell{100.00} \\
\evalmodel{claude}{Claude-Sonnet-4.6} & \acccell{100.00} & \acccell{100.00} & \acccell{100.00} & \acccell{100.00} & \acccell{100.00} & \acccell{100.00} \\
\addlinespace[2pt]
\evalmodel{gemini}{Gemini-3.7-Flash}$^{*}$ & \acccell{100.00} & \acccell{100.00} & \acccell{100.00} & \acccell{100.00} & \acccell{100.00} & \acccell{100.00} \\
\evalmodel{gemini}{Gemini-3.6-Flash} & \acccell{100.00} & \acccell{100.00} & \acccell{100.00} & \acccell{100.00} & \acccell{100.00} & \acccell{100.00} \\
\evalmodel{gemini}{Gemini-3.5-Flash-Lite} & \acccell{99.90} & \acccell{100.00} & \acccell{100.00} & \acccell{100.00} & \acccell{100.00} & \acccell{100.00} \\
\addlinespace[2pt]
\evalmodel{deepseek}{DeepSeek-V4-Flash}$^{*}$ & \acccell{64.75} & \acccell{98.50} & \acccell{99.80} & \acccell{99.90} & \acccell{100.00} & \acccell{100.00} \\
\evalmodel{deepseek}{DeepSeek-V4-Pro} & \acccell{84.90} & \acccell{94.20} & \acccell{98.30} & \acccell{99.70} & \acccell{99.90} & \acccell{100.00} \\
\addlinespace[2pt]
\evalmodel{qwen}{Qwen-3.8-Flash}$^{*}$ & \acccell{98.50} & \acccell{100.00} & \acccell{100.00} & \acccell{100.00} & \acccell{100.00} & \acccell{100.00} \\
\evalmodel{qwen}{Qwen-3.8-Max} & \acccell{95.30} & \acccell{99.80} & \acccell{99.80} & \acccell{100.00} & \acccell{100.00} & \acccell{100.00} \\
\evalmodel{qwen}{Qwen-3.7-Flash} & \acccell{98.40} & \acccell{100.00} & \acccell{100.00} & \acccell{100.00} & \acccell{100.00} & \acccell{100.00} \\
\evalmodel{qwen}{Qwen-3.7-Max} & \acccell{94.10} & \acccell{100.00} & \acccell{100.00} & \acccell{100.00} & \acccell{100.00} & \acccell{100.00} \\
\addlinespace[2pt]
\evalmodel{meta}{Muse-Spark-1.1} & \acccell{92.30} & \acccell{100.00} & \acccell{100.00} & \acccell{100.00} & \acccell{100.00} & \acccell{100.00} \\
\evalmodel{meta}{Llama-4-Maverick} & \acccell{100.00} & \acccell{100.00} & \acccell{100.00} & \acccell{100.00} & \acccell{100.00} & \acccell{100.00} \\
\addlinespace[2pt]
\evalmodel{grok}{Grok-4.6} & \acccell{67.20} & \acccell{95.00} & \acccell{96.70} & \acccell{98.80} & \acccell{99.70} & \acccell{99.80} \\
\evalmodel{grok}{Grok-4.5} & \acccell{77.80} & \acccell{93.40} & \acccell{97.10} & \acccell{99.40} & \acccell{99.80} & \acccell{99.90} \\
\evalmodel{grok}{Grok-4.3} & \acccell{97.90} & \acccell{99.80} & \acccell{100.00} & \acccell{100.00} & \acccell{100.00} & \acccell{100.00} \\
\addlinespace[2pt]
\evalmodel{zhipu}{GLM-5.3} & \acccell{68.00} & \acccell{97.60} & \acccell{99.60} & \acccell{99.70} & \acccell{99.90} & \acccell{100.00} \\
\evalmodel{zhipu}{GLM-5.3-Flash} & \acccell{97.40} & \acccell{99.70} & \acccell{99.90} & \acccell{100.00} & \acccell{100.00} & \acccell{100.00} \\
\evalmodel{zhipu}{GLM-5.2} & \acccell{90.90} & \acccell{98.60} & \acccell{99.90} & \acccell{100.00} & \acccell{100.00} & \acccell{100.00} \\
\addlinespace[2pt]
\evalmodel{moonshot}{Kimi-K2.6}$^{*}$ & \acccell{96.60} & \acccell{100.00} & \acccell{100.00} & \acccell{100.00} & \acccell{100.00} & \acccell{100.00} \\
\addlinespace[2pt]
\evalmodel{minimax}{MiniMax-M3}$^{*}$ & \acccell{77.50} & \acccell{99.40} & \acccell{100.00} & \acccell{100.00} & \acccell{100.00} & \acccell{100.00} \\
\midrule
\textbf{Mean (all)} & \textbf{\acccell{91.65}} & \textbf{\acccell{99.03}} & \textbf{\acccell{99.64}} & \textbf{\acccell{99.90}} & \textbf{\acccell{99.97}} & \textbf{\acccell{99.99}} \\
\quad 7 representative models & \acccell{91.05} & \acccell{99.70} & \acccell{99.97} & \acccell{99.99} & \acccell{100.00} & \acccell{100.00} \\
\quad 18 additional models & \acccell{91.88} & \acccell{98.77} & \acccell{99.52} & \acccell{99.87} & \acccell{99.96} & \acccell{99.98} \\
\textbf{Worst-model (all)} & \textbf{\acccell{64.75}} & \textbf{\acccell{93.40}} & \textbf{\acccell{96.70}} & \textbf{\acccell{98.80}} & \textbf{\acccell{99.70}} & \textbf{\acccell{99.80}} \\
Models $\geq95\%$ & 15/25 & 23/25 & 25/25 & 25/25 & 25/25 & 25/25 \\
Models $\geq99\%$ & 9/25 & 19/25 & 22/25 & 24/25 & 25/25 & 25/25 \\
\bottomrule
\end{tabular*}
\par\smallskip
\parbox{\linewidth}{\footnotesize $^\ast$ marks the seven models used for probe construction.}
\end{table}

\subsection{Effectiveness of Probe Construction}
\label{sec:eval:effectiveness}

% 275 candidate questions
% Top/Middle/Bottom 5
% Baseline and optimized ranking effectiveness
% Worst-model comparison
% Table: eval-ranking
\bheading{For Challenge 1: Question selection.}
We first evaluate whether the diversity score helps identify questions with greater identification potential.
We rank all 275 candidate questions by their diversity scores and select five questions each from the top, middle, and bottom of the ranking.
Table~\ref{tab:eval-ranking} compares their accuracy before (Baseline) and after (Optimized) option-set optimization.
Notably, all Bottom 5 questions have a diversity score of zero, as all seven models generate the same valid options for these questions.

For single-probe identification, the Top 5 achieve a mean accuracy of \accp{47.01}, outperforming the Middle 5 (\accp{41.08}) and Bottom 5 (\accp{29.85}).
The differences become even larger after option-set optimization, with the Top 5 achieving \accp{73.40}, compared with \accp{54.97} and \accp{39.57} for the other two groups.
The same trend holds for five-probe joint identification: higher-ranked questions achieve better mean accuracy. The worst-model accuracy also follows this trend, reaching \accp{86.25}, \accp{80.30}, and \accp{58.85} for the Top 5, Middle 5, and Bottom 5, respectively.

\finding{
Higher-ranked questions achieve better identification accuracy, both before and after option-set optimization, demonstrating the effectiveness of diversity scores in Stage 1.
}

% All 30 selected probes
% Baseline vs. optimized
% Single-probe accuracy
% Identification accuracy vs. number of probes
% Example and optimization metrics
\bheading{For Challenge 2: Option-set optimization.}
We evaluate option-set optimization on all 30 probes using the seven representative models.
The smallest initial option set contains 22 options, yielding over 4.19 million possible subsets, but we allow at most 12 optimization rounds per probe.

\begin{figure}[t]
    \centering
    \begin{minipage}[c]{1.68in}
        \centering
        \includegraphics[scale=1]{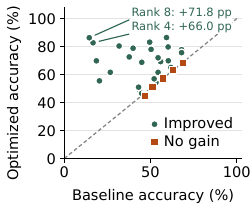}
        \par\smallskip
        {\footnotesize (a) Single-probe accuracy}
    \end{minipage}%
    \hfill%
    \begin{minipage}[c]{1.68in}
        \centering
        \includegraphics[scale=1]{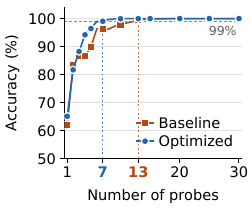}
        \par\smallskip
        {\footnotesize (b) Joint identification accuracy}
    \end{minipage}
    \caption{
        Effect of option-set optimization on single-probe (a) and joint (b) identification. 
        Optimization reduces the probes needed for 99\% mean accuracy from 13 to 7.
    }
\label{fig:eval-opt}
\end{figure}

As shown in Figure~\ref{fig:eval-opt}(a), optimization improves most probes; 25 probes improve, four remain unchanged, and one shows a slight decrease.
The mean single-probe accuracy increases from \accp{47.23} to \accp{67.24}.
For joint identification, the mean accuracy with the top five probes increases from \accp{89.79} to \accp{96.44}.
Figure~\ref{fig:eval-opt}(b) shows that optimization reduces the number of probes needed for 99\% mean accuracy from 13 to 7.
For 99\% worst-model accuracy, the required number of probes decreases from 16 to 9.

Notably, even high-ranking questions may have low initial accuracy and require option-set optimization.
For example, the fourth-ranked question has a diversity score of 0.7692, but its single-probe accuracy increases from \accp{16.67} to \accp{82.70} after optimization.
Similarly, the eighth-ranked question improves from \accp{14.50} to \accp{86.29}.

The four unchanged probes already have relatively high initial accuracies (\accp{51.16}--\accp{68.57}) and trigger early stopping after six rounds without improvement.
For the only probe with a decrease, the optimizer retains the updated option set because both $MI_{\mathrm{ratio}}$ and $A_p$ improve slightly, despite a small drop in actual accuracy (\accp{46.74} to \accp{45.01}).

The optimization metrics further support these results.
Across all 30 probes, the mean $MI_{\mathrm{ratio}}$ increases from 0.4168 to 0.6492, and $A_p$ increases from 0.4925 to 0.6933, while $R_2$ remains relatively stable (0.6189 to 0.6077).

\bheading{Our LLM-guided vs. rule-based optimization.}
To evaluate the contribution of LLM guidance, we replace the LLM optimizer with a rule-based method and compare their performance. LLM-guided optimization improves mean single-probe accuracy by 18.02 percentage points and five-probe worst-model accuracy by 11.90 percentage points over the rule-based method. Details are provided in Appendix~\ref{app:opt-comparison}.

\finding{
Stage 2's LLM-guided optimization significantly improves probe accuracy and outperforms the rule-based method. 
Moreover, high-ranking questions may still require option-set optimization to achieve high identification accuracy.
}

\begin{table}[t]
\centering
\caption{Identification accuracy (\%) for questions ranked by diversity score. Higher-ranked groups perform better before and after option-set optimization.}
\label{tab:eval-ranking}
\footnotesize
\setlength{\tabcolsep}{2pt}
\renewcommand{\arraystretch}{1.15}
\begin{tabular*}{\linewidth}{@{\extracolsep{\fill}}lcrrrr@{}}
\toprule
\multirow{2}{*}[-3pt]{\makecell[c]{\textbf{Question}\\\textbf{rank}}}
 & \multirow{2}{*}[-3pt]{\makecell{\textbf{Diversity}\\\textbf{score}}}
 & \multicolumn{2}{c}{\textbf{Single-probe mean}}
& \multicolumn{2}{c}{\textbf{Five-probe joint}} \\
\cmidrule(lr){3-4}\cmidrule(l){5-6}
 & & \textbf{Baseline} & \textbf{Optimized} & \textbf{Baseline} & \textbf{Optimized} \\
\midrule
Top 5    & 0.7863 & \acc{47.01} & \textbf{\acc{73.40}} & \acc{89.79} & \textbf{\acc{96.44}} \\
Middle 5 & 0.5107 & \acc{41.08} & \textbf{\acc{54.97}} & \acc{89.51} & \textbf{\acc{94.64}} \\
Bottom 5 & 0.0000 & \acc{29.85} & \textbf{\acc{39.57}} & \acc{73.61} & \textbf{\acc{84.96}} \\
\bottomrule
\end{tabular*}
\end{table}

% \begin{table}[t]
% \centering
% \caption{Effectiveness of question selection and option-set
% optimization (accuracy in percent).}
% \label{tab:eval-ranking}
% \footnotesize
% \setlength{\tabcolsep}{3pt}
% \renewcommand{\arraystretch}{1.12}
% \resizebox{\columnwidth}{!}{%
% \begin{tabular}{@{}llcrrrr@{}}
% \toprule
%  & & Diversity
%  & \multicolumn{2}{c}{Single probe}
%  & \multicolumn{2}{c}{Five probes} \\
% \cmidrule(lr){4-5}\cmidrule(l){6-7}
% Question rank & Option set & score
%  & Mean & Worst & Mean & Worst \\
% \midrule
% \multirow{2}{*}{Top 5 (1--5)}
%  & Baseline  & \multirow{2}{*}{0.7863}
%  & 47.01 & 19.85 & 89.79 & 66.60 \\
%  & Optimized &
%  & \textbf{73.40} & \textbf{31.14}
%  & \textbf{96.44} & \textbf{86.25} \\
% \midrule
% \multirow{2}{*}{Middle 5 (136--140)}
%  & Baseline & \multirow{2}{*}{0.5107}
%  & 41.08 & 22.94 & 89.51 & 73.60 \\
%  & Optimized &
%  & \textbf{54.97} & \textbf{26.48}
%  & \textbf{94.64} & \textbf{80.30} \\
% \midrule
% \multirow{2}{*}{Bottom 5 (271--275)}
%  & Baseline & \multirow{2}{*}{0.0000}
%  & 29.85 & \textbf{23.22} & 73.61 & 12.30 \\
%  & Optimized &
%  & \textbf{39.57} & 8.03
%  & \textbf{84.96} & \textbf{58.85} \\
% \bottomrule
% \end{tabular}%
% }
% \end{table}

% Source: results/07-ranking-ablation/formal-01/analysis/ranking-ablation.pdf
% \begin{figure*}[tp]
% \centering
% \includegraphics[width=\linewidth]{figures/eval-ranking.pdf}
% \caption{Baseline and optimized options for the three fixed ranking groups. Single-probe averages and five-probe joint accuracy are evaluated separately.}
% \label{fig:eval-ranking}
% \end{figure*}

\begin{table*}[t]
\centering
\caption{Identification accuracy (\%) across four agents and seven LLMs using the top 10 probes.}
\label{tab:eval-agent}
\footnotesize
\setlength{\tabcolsep}{1pt}
\renewcommand{\arraystretch}{1.15}
\begin{tabular*}{\linewidth}{@{}l@{\extracolsep{\fill}\hspace{6pt}}>{\hspace{3pt}}r<{\hspace{3pt}}@{\extracolsep{0pt}\hspace{2pt}}c@{\hspace{2pt}}>{\hspace{3pt}}l<{\hspace{3pt}}@{\extracolsep{\fill}\hspace{6pt}}>{\hspace{3pt}}r<{\hspace{3pt}}@{\extracolsep{0pt}\hspace{2pt}}c@{\hspace{2pt}}>{\hspace{3pt}}l<{\hspace{3pt}}@{\extracolsep{\fill}\hspace{6pt}}>{\hspace{3pt}}r<{\hspace{3pt}}@{\extracolsep{0pt}\hspace{2pt}}c@{\hspace{2pt}}>{\hspace{3pt}}l<{\hspace{3pt}}@{\extracolsep{\fill}\hspace{6pt}}>{\hspace{3pt}}r<{\hspace{3pt}}@{\extracolsep{0pt}\hspace{2pt}}c@{\hspace{2pt}}>{\hspace{3pt}}l<{\hspace{3pt}}@{\extracolsep{\fill}\hspace{6pt}}>{\hspace{3pt}}r<{\hspace{3pt}}@{\extracolsep{0pt}\hspace{2pt}}c@{\hspace{2pt}}>{\hspace{3pt}}l<{\hspace{3pt}}@{\extracolsep{\fill}\hspace{6pt}}>{\hspace{3pt}}r<{\hspace{3pt}}@{\extracolsep{0pt}\hspace{2pt}}c@{\hspace{2pt}}>{\hspace{3pt}}l<{\hspace{3pt}}@{}}
\toprule
\textbf{Model}
    & \multicolumn{3}{c}{\agentIcon{\includegraphics[height=1.2em]{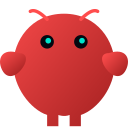}}{\textbf{OpenClaw}}}
    & \multicolumn{3}{c}{\agentIcon{\includegraphics[height=1.2em]{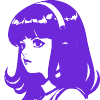}}{\textbf{Hermes}}}
    & \multicolumn{3}{c}{\agentIcon{\includegraphics[height=1.2em]{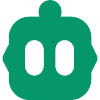}}{\textbf{Cline}}}
    & \multicolumn{3}{c}{\agentIcon{\includegraphics[height=1.2em]{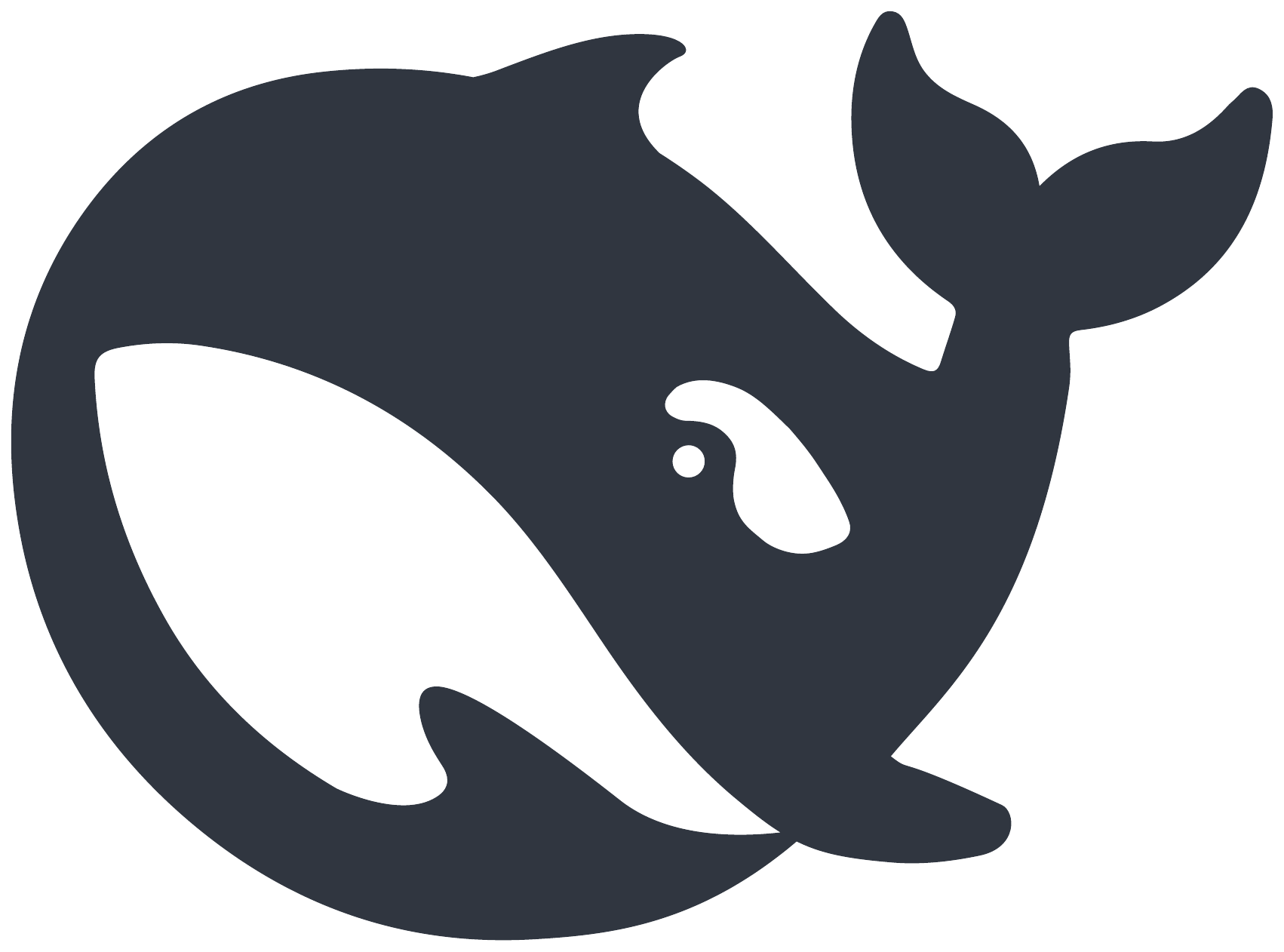}}{\textbf{Harness}}}
    & \multicolumn{3}{c}{\textbf{Mean}}
    & \multicolumn{3}{c}{\textbf{Min.}} \\
\midrule
\evalmodel{openai}{GPT-5.6-Luna}
    & \cellcolor{acc95}\acc{98.50} & $\to$ & \cellcolor{acc100}\acc{100.00}
    & \cellcolor{acc100}\acc{100.00} & $\to$ & \cellcolor{acc100}\acc{100.00}
    & \cellcolor{acc100}\acc{100.00} & $\to$ & \cellcolor{acc100}\acc{100.00}
    & \cellcolor{acc99}\acc{99.70} & $\to$ & \cellcolor{acc100}\acc{100.00}
    & \cellcolor{acc99}\acc{99.55} & $\to$ & \cellcolor{acc100}\acc{100.00}
    & \cellcolor{acc95}\acc{98.50} & $\to$ & \cellcolor{acc100}\acc{100.00} \\
\evalmodel{claude}{Claude-Sonnet-5}
    & \cellcolor{accBelow70}\acc{67.10} & $\to$ & \cellcolor{acc95}\acc{96.50}
    & \cellcolor{acc100}\acc{100.00} & $\to$ & \cellcolor{acc100}\acc{100.00}
    & \cellcolor{acc90}\acc{90.00} & $\to$ & \cellcolor{acc100}\acc{100.00}
    & \cellcolor{acc95}\acc{97.40} & $\to$ & \cellcolor{acc100}\acc{100.00}
    & \cellcolor{acc80}\acc{88.63} & $\to$ & \cellcolor{acc99}\acc{99.13}
    & \cellcolor{accBelow70}\acc{67.10} & $\to$ & \cellcolor{acc95}\acc{96.50} \\
\evalmodel{gemini}{Gemini-3.7-Flash}
    & \cellcolor{acc100}\acc{100.00} & $\to$ & \cellcolor{acc100}\acc{100.00}
    & \cellcolor{acc100}\acc{100.00} & $\to$ & \cellcolor{acc100}\acc{100.00}
    & \cellcolor{acc100}\acc{100.00} & $\to$ & \cellcolor{acc100}\acc{100.00}
    & \cellcolor{acc100}\acc{100.00} & $\to$ & \cellcolor{acc100}\acc{100.00}
    & \cellcolor{acc100}\acc{100.00} & $\to$ & \cellcolor{acc100}\acc{100.00}
    & \cellcolor{acc100}\acc{100.00} & $\to$ & \cellcolor{acc100}\acc{100.00} \\
\evalmodel{deepseek}{DeepSeek-V4-Flash}
    & \cellcolor{acc80}\acc{81.20} & $\to$ & \cellcolor{acc99}\acc{99.80}
    & \cellcolor{acc95}\acc{97.00} & $\to$ & \cellcolor{acc100}\acc{100.00}
    & \cellcolor{acc95}\acc{97.20} & $\to$ & \cellcolor{acc100}\acc{100.00}
    & \cellcolor{acc95}\acc{97.80} & $\to$ & \cellcolor{acc100}\acc{100.00}
    & \cellcolor{acc90}\acc{93.30} & $\to$ & \cellcolor{acc99}\acc{99.95}
    & \cellcolor{acc80}\acc{81.20} & $\to$ & \cellcolor{acc99}\acc{99.80} \\
\evalmodel{qwen}{Qwen-3.8-Flash}
    & \cellcolor{acc70}\acc{78.30} & $\to$ & \cellcolor{acc99}\acc{99.40}
    & \cellcolor{acc80}\acc{88.50} & $\to$ & \cellcolor{acc100}\acc{100.00}
    & \cellcolor{acc70}\acc{78.30} & $\to$ & \cellcolor{acc99}\acc{99.10}
    & \cellcolor{acc70}\acc{75.20} & $\to$ & \cellcolor{acc95}\acc{97.70}
    & \cellcolor{acc80}\acc{80.08} & $\to$ & \cellcolor{acc99}\acc{99.05}
    & \cellcolor{acc70}\acc{75.20} & $\to$ & \cellcolor{acc95}\acc{97.70} \\
\evalmodel{moonshot}{Kimi-K2.6}
    & \cellcolor{acc95}\acc{98.80} & $\to$ & \cellcolor{acc100}\acc{100.00}
    & \cellcolor{acc99}\acc{99.10} & $\to$ & \cellcolor{acc100}\acc{100.00}
    & \cellcolor{acc100}\acc{100.00} & $\to$ & \cellcolor{acc100}\acc{100.00}
    & \cellcolor{acc100}\acc{100.00} & $\to$ & \cellcolor{acc100}\acc{100.00}
    & \cellcolor{acc99}\acc{99.48} & $\to$ & \cellcolor{acc100}\acc{100.00}
    & \cellcolor{acc95}\acc{98.80} & $\to$ & \cellcolor{acc100}\acc{100.00} \\
\evalmodel{minimax}{MiniMax-M3}
    & \cellcolor{acc99}\acc{99.90} & $\to$ & \cellcolor{acc100}\acc{100.00}
    & \cellcolor{acc95}\acc{95.50} & $\to$ & \cellcolor{acc100}\acc{100.00}
    & \cellcolor{acc99}\acc{99.40} & $\to$ & \cellcolor{acc100}\acc{100.00}
    & \cellcolor{acc95}\acc{97.00} & $\to$ & \cellcolor{acc100}\acc{100.00}
    & \cellcolor{acc95}\acc{97.95} & $\to$ & \cellcolor{acc100}\acc{100.00}
    & \cellcolor{acc95}\acc{95.50} & $\to$ & \cellcolor{acc100}\acc{100.00} \\
\midrule
\textbf{Mean}
    & \cellcolor{acc80}\acc{89.11} & $\to$ & \cellcolor{acc99}\acc{99.39}
    & \cellcolor{acc95}\acc{97.16} & $\to$ & \cellcolor{acc100}\acc{100.00}
    & \cellcolor{acc90}\acc{94.99} & $\to$ & \cellcolor{acc99}\acc{99.87}
    & \cellcolor{acc95}\acc{95.30} & $\to$ & \cellcolor{acc99}\acc{99.67}
    & \cellcolor{acc90}\acc{94.14} & $\to$ & \cellcolor{acc99}\acc{99.73} & \multicolumn{3}{c}{---} \\
\textbf{Min.}
    & \cellcolor{accBelow70}\acc{67.10} & $\to$ & \cellcolor{acc95}\acc{96.50}
    & \cellcolor{acc80}\acc{88.50} & $\to$ & \cellcolor{acc100}\acc{100.00}
    & \cellcolor{acc70}\acc{78.30} & $\to$ & \cellcolor{acc99}\acc{99.10}
    & \cellcolor{acc70}\acc{75.20} & $\to$ & \cellcolor{acc95}\acc{97.70}
    & \multicolumn{3}{c}{---}
    & \cellcolor{accBelow70}\acc{67.10} & $\to$ & \cellcolor{acc95}\acc{96.50} \\
\bottomrule
\end{tabular*}

\par\smallskip
\parbox{\linewidth}{\footnotesize
Each cell shows accuracy with 10 $\rightarrow$ 30 total responses
(one versus three responses per probe).
}
\end{table*}

\subsection{Evaluation with Real-World Agents}
\label{sec:eval:agent}

We evaluate CAPTA with four agents and seven representative LLMs, resulting in 28 agent--LLM combinations.
Each agent--LLM pair performs a weather query and completes CAPTA verification through the weather-service skill in an independent session.
Agents access the service through their normal tool interfaces, without exposing their prompts or execution traces to the service.

During runtime, 
we collect five responses per probe for each pair and use the same reference fingerprints across all four agents.
To evaluate different response budgets, we randomly sample 1, 2, 3, or 4 responses per probe without replacement, yielding 10, 20, 30, or 40 responses, respectively.
We repeat this sampling process 1,000 times and report the mean accuracy.
With 10 probes per interaction, these budgets correspond to 1--4 agent--service interactions.

\bheading{Identification in real-world agents.}
All four agents successfully complete CAPTA verification without refusing any probes.
Table~\ref{tab:eval-agent} shows the identification accuracy across 28 agent--LLM pairs.
With 10 responses (one interaction), CAPTA achieves a mean accuracy of \accp{94.14}, with three agents achieving approximately \accp{95.0} accuracy or higher.
With 30 responses (three interactions), the mean accuracy increases to \accp{99.73}, and the worst-case accuracy reaches \accp{96.50}.
With 40 responses (four interactions), all 28 pairs achieve \accp{100} accuracy.

\bheading{Effects of agent contexts.}
Identification accuracy varies across agents, with OpenClaw performing relatively worse.
Three models show lower accuracy with OpenClaw, especially Claude-Sonnet-5, which achieves only \accp{67.10}.
We find that OpenClaw adds a security notice stating that external content is untrusted.
This notice may change LLM preferences and lead to different responses.

Although we collect fingerprints using \texttt{web\_fetch}, agents use their own tools to submit answers at runtime.
OpenClaw and Harness use \texttt{web\_fetch}, Hermes uses \texttt{curl}, and Cline uses \texttt{fetch\_web\_content}.
Across the four agents, Gemini-3.7-Flash achieves \accp{100} accuracy on all agents.
GPT-5.6-Luna, Kimi-K2.6, and MiniMax-M3 also perform well.
This shows that CAPTA works with different agent tool interfaces.

% Across the four agents, Gemini-3.7-Flash achieves \accp{100} accuracy on all agents.
% GPT-5.6-Luna, Kimi-K2.6, and MiniMax-M3 also perform well.
% In contrast, Qwen-3.8-Flash and Claude-Sonnet-5 are more affected by agent contexts, with mean accuracies of \accp{80.08} and \accp{88.63}, respectively, using 10 responses.

\finding{
CAPTA works effectively in real-world agents, achieving 99.73\% mean identification accuracy across 28 agent--LLM pairs with only three interactions.
Agent contexts may cause preference shifts, but additional responses can mitigate their impact.
}

\subsection{Cross-Task Identification}
\label{sec:eval:cross-task}

We use seven representative LLMs and the top ten probes to evaluate CAPTA's ability to reuse reference fingerprints across different tasks. 
We consider two common tasks, Weather and Flight. 
We first collect shared reference fingerprints in a neutral resource-query task without task-specific information, and then reuse them for identification in the two task-specific settings.

\bheading{Identification across tasks.}
As shown in Table~\ref{tab:eval-scenario}, CAPTA achieves \accp{99.75} mean accuracy in the neutral task. Using the same reference fingerprints, it achieves \accp{99.09} in Weather and \accp{98.88} in Flight, with decreases of only 0.66 and 0.87 percentage points, respectively. The worst-model accuracy remains above \accp{94} in both tasks. 

Task contexts may change model preferences on individual probes. For example, for a museum-exhibit selection probe, MiniMax-M3 most frequently selects \textit{dinosaur} in the neutral context (24/51 responses), but selects space-related options in nine out of ten responses under the Flight task. Nevertheless, combining multiple probes provides complementary evidence and maintains high identification accuracy despite these changes. Moreover, since the service typically knows the task it provides, it can select probes unrelated to that task to reduce the influence of task-specific contexts.

\finding{
CAPTA can reuse the same probes and fingerprints across different tasks. Although task contexts may change model preferences on some probes, combining multiple probes still enables accurate identification.
}

\begin{table}[t]
\centering
\caption{Identification accuracy (\%) across task contexts.}
\label{tab:eval-scenario}
\footnotesize
\setlength{\tabcolsep}{2pt}
\renewcommand{\arraystretch}{1.15}
\begin{tabular*}{\linewidth}{@{\extracolsep{\fill}}lccc@{}}
\toprule
\textbf{Model} & \textbf{Neutral task} & \textbf{Weather task} & \textbf{Flight task} \\
\midrule
\evalmodel{openai}{GPT-5.6-Luna}
    & \acccell{100.00} & \acccell{99.79} & \acccell{99.98} \\
\evalmodel{claude}{Claude-Sonnet-5}
    & \acccell{100.00} & \acccell{100.00} & \acccell{100.00} \\
\evalmodel{gemini}{Gemini-3.7-Flash}
    & \acccell{100.00} & \acccell{100.00} & \acccell{99.97} \\
\evalmodel{deepseek}{DeepSeek-V4-Flash}
    & \acccell{98.73} & \acccell{94.09} & \acccell{98.16} \\
\evalmodel{qwen}{Qwen-3.8-Flash}
    & \acccell{100.00} & \acccell{100.00} & \acccell{100.00} \\
\evalmodel{moonshot}{Kimi-K2.6}
    & \acccell{100.00} & \acccell{100.00} & \acccell{99.94} \\
\evalmodel{minimax}{MiniMax-M3}
    & \acccell{99.52} & \acccell{99.76} & \acccell{94.12} \\
\midrule
\textbf{Mean}
    & \textbf{\acccell{99.75}}
    & \textbf{\acccell{99.09}}\,\downchange{\formatdecimal{0.66}}
    & \textbf{\acccell{98.88}}\,\downchange{\formatdecimal{0.87}} \\
\textbf{Worst-model}
    & \textbf{\acccell{98.73}}
    & \textbf{\acccell{94.09}}\,\downchange{\formatdecimal{4.64}}
    & \textbf{\acccell{94.12}}\,\downchange{\formatdecimal{4.61}} \\
\bottomrule
\end{tabular*}

\par\smallskip
\parbox{\linewidth}{\footnotesize
Arrows indicate decreases in percentage points relative to the neutral task.
}

\end{table}

\subsection{Open-Set Identification}
\label{sec:eval:open-set}

We evaluate CAPTA in an open-set setting, where the target model may not belong to the service's current candidate set.
In practice, a service may maintain fingerprints for a set of known models, while agents using other models may also access the service. CAPTA must therefore either correctly identify a known model or reject an unknown one.

To evaluate this capability, we use the 25 models and randomly divide them into known and unknown groups.
Known models retain their reference fingerprints and form the candidate set for identification, whereas fingerprints of unknown models are excluded from the candidate set.
We evaluate different candidate-set sizes by randomly selecting 5, 10, 15, or 20 models as unknown,
with the remaining models treated as known. 
We also include an all-known setting (0 unknown, 25 known) as a reference.

Following existing open-set recognition methods~\cite{DBLP:conf/iclr/Vaze0VZ22, DBLP:conf/eccv/CruzRGB24}, we introduce a threshold-based rejection mechanism with target false-rejection rates of 1\%, 5\%, and 10\%.
For each known model--probe pair, we use 40 of the 50 sampled responses to construct fingerprints and the
remaining 10 to calibrate the rejection threshold.

\bheading{Results across different U/K settings.}
As shown in Table~\ref{tab:eval-open-set}, CAPTA detects most unknown models, with higher detection rates when fewer known models remain.
With 20 unknown and 5 known models, the detection rate is at least \accp{95.47} across all three target false-rejection rates.
Even with 5 unknown and 20 known models, CAPTA achieves \accp{89.86} unknown-model detection and \accp{94.73} known-model identification accuracy at a 5\% target false-rejection rate.

\bheading{Trade-off between unknown-model detection and known-model identification.}
Higher false-rejection rates can improve unknown-model detection but reduce known-model identification accuracy.
For example, with 20 known models, increasing the target false-rejection rate from 1\% to 10\% raises the unknown-model detection rate from \accp{79.74} to \accp{94.50}, while reducing known-model identification accuracy from \accp{98.59} to \accp{89.78}.

% \bheading{Effect of similar models.}
% Similar preferences between known and unknown models can also affect detection.
% When Grok-4.5 is unknown but Grok-4.6 is known, the detection rate for Grok-4.5 is only \accp{7.83}.
% Similarly, the detection rate for Grok-4.6 is only \accp{17.07} when Grok-4.5 remains known.
% When both models are treated as unknown, their detection rates increase to \accp{100.00} and \accp{97.30}, respectively.

\finding{
CAPTA can detect models excluded from the candidate set while maintaining high identification accuracy for known models. 
Higher false-rejection rates can improve unknown-model detection but reduce known-model identification accuracy.
}

\subsection{Practical Considerations}
\label{sec:eval:practical}

This section evaluates practical considerations using the seven representative models, including multi-probe submission, sampling budget, and temperature. 
We finally analyze the token costs and report confidence intervals.

\bheading{Multi-probe submission within a single verification.}
In practice, a service can present multiple probes within a single verification process. We consider two approaches: providing a group of probes in a single request (Batch), and presenting probes one by one while retaining previous questions and answers in the context (Cumulative). 
Our results show that both approaches achieve high identification accuracy. With 10 probes per group, Batch achieves \accp{99.73} mean accuracy after two interactions, while Cumulative achieves \accp{100.0} accuracy across all models after three interactions. More details are provided in Appendix~\ref{app:multi-probe}.

\begin{figure}[t]
    \centering
    \begin{minipage}[c]{1.68in}
        \centering
        \includegraphics[scale=1]{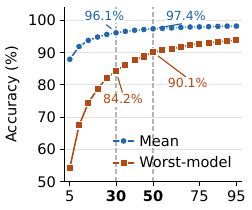}
        \par\smallskip
        {\footnotesize (a) Responses per model and probe}
    \end{minipage}%
    \hfill%
    \begin{minipage}[c]{1.68in}
        \centering
        \includegraphics[scale=1]{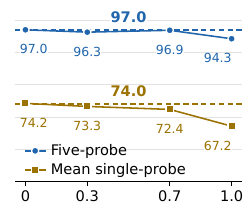}
        \par\smallskip
        {\footnotesize (b) Runtime temperature}
    \end{minipage}
    \caption{Identification accuracy using the top five probes with
different fingerprint sample sizes (a) and runtime temperatures (b).
Dashed lines in (b) show accuracy at the default temperature.}
    \label{fig:eval-practical}
\end{figure}

\bheading{Sampling budget.}
As shown in Figure~\ref{fig:eval-practical}(a), identification accuracy improves with more fingerprint responses, although the gains diminish beyond 50.
Mean accuracy rises from \accp{96.11} with 30 responses to \accp{97.37} with 50 and \accp{98.20} with 95.
Therefore, we use 30 responses for Stage 2 optimization and 50 for Stage 3 fingerprint construction to balance cost and accuracy.

\bheading{Temperature.}
We collect reference fingerprints at the default temperature and test identification at different runtime temperatures.
As GPT-5.6-Luna and Claude-Sonnet-5 do not support temperature control, we evaluate the remaining five models.
As shown in Figure~\ref{fig:eval-practical}(b), temperature changes have a limited impact on identification accuracy. From 0 to 0.7, both mean single-probe and five-probe joint accuracy remain close to their default-temperature results. At temperature 1.0, mean single-probe accuracy drops by 6.8 percentage points, while five-probe joint accuracy decreases by only 2.7 percentage points.

\bheading{Token cost.}
During preparation, CAPTA consumes 76.26M tokens in total. Based on the highest API prices among the tested models (\$5 / \$25 per million input / output tokens), the estimated cost is \$485.
Specifically, Stage 1 consumes 3.30M tokens for question and option generation, Stage 2 consumes 51.22M tokens for option-set optimization, 
and Stage 3 consumes 21.74M tokens for fingerprint collection.

The optimized probes can be reused for new models. 
Adding a new model only requires collecting 50 responses for each of the 30 probes.
Based on the models we evaluated, this requires approximately 0.87--1.64M tokens, at an estimated cost of \$4.5--16.9.

During runtime, each probe consumes approximately 700 tokens on average across the 25 models. At the same API prices, using 10--30 probes incurs an estimated additional cost of only \$0.046--0.14 per identification. 
This estimate covers only the additional cost of answering CAPTA probes; 
the actual runtime cost also depends on the agent's context, including system prompts and conversation history.

\bheading{Confidence intervals.}
For the model identification evaluation (Section~\ref{sec:eval:identification}), we report 95\% confidence intervals for mean identification accuracy in Appendix~\ref{app:confidence}.
The intervals are narrow across all probe budgets.
For example, the interval is only 0.244 percentage points wide with 10 probes, indicating stable identification accuracy.

\begin{table}[t]
\centering
\caption{Unknown-model detection and known-model identification (\%) using 30 probes at different target false-rejection rates.}
\label{tab:eval-open-set}
\footnotesize
\setlength{\tabcolsep}{2pt}
\renewcommand{\arraystretch}{1.15}
\begin{tabular*}{\linewidth}{@{\extracolsep{\fill}}lrrrrrr@{}}
\toprule
 & \multicolumn{2}{c}{\textbf{1\%}}
 & \multicolumn{2}{c}{\textbf{5\%}}
 & \multicolumn{2}{c}{\textbf{10\%}} \\
\cmidrule(lr){2-3}\cmidrule(lr){4-5}\cmidrule(l){6-7}
{\boldmath$U/K$} & \textbf{Unknown} & \textbf{Known} & \textbf{Unknown} & \textbf{Known} & \textbf{Unknown} & \textbf{Known} \\
\midrule
20/5 & \acccell{95.47} & \acccell{98.34} & \acccell{97.94} & \acccell{94.96} & \acccell{98.84} & \acccell{89.00} \\
15/10 & \acccell{90.97} & \acccell{98.06} & \acccell{95.31} & \acccell{94.27} & \acccell{97.35} & \acccell{89.51} \\
10/15 & \acccell{83.29} & \acccell{98.77} & \acccell{92.50} & \acccell{94.97} & \acccell{96.02} & \acccell{89.63} \\
5/20 & \acccell{79.74} & \acccell{98.59} & \acccell{89.86} & \acccell{94.73} & \acccell{94.50} & \acccell{89.78} \\
\midrule
0/25 & -- & \acccell{98.87} & -- & \acccell{94.95} & -- & \acccell{89.83} \\
\bottomrule
\end{tabular*}
\par\smallskip
\parbox{\linewidth}{\footnotesize
    $U/K$: number of unknown/known models.
    Unknown: correct rejection rate of unknown models. 
    Known: correct identification rate of known models.
}
\end{table}

\section{Discussion}
\label{sec:discussion}

% This section discusses the practical deployment, robustness, security implications of CAPTA, and its limitations.

\subsection{Practical Deployment of CAPTA}

\bheading{Agent--service interactions.}
Agents can access external services in different ways. 
When accessing services through APIs, agents may require user approval to proceed with verification. In our tests, agents successfully complete CAPTA verification and resume their tasks once users approve it.

Besides, many widely used services, such as Notion and Gmail, provide Skills or Plugins for agent integration. We package the Weather service as a Skill and indicate that verification may be required. Across all 28 agent--LLM combinations, agents automatically complete CAPTA verification without user approval. Moreover, none of the agents refuses verification or issues security warnings.

These results demonstrate that CAPTA supports both direct API access and common Skill-based service integrations, with the latter enabling more seamless verification.

\bheading{Users or LLMs deliberately refuse CAPTA.}
CAPTA uses the existing CAPTCHA verification process, so users or agents are unlikely to notice its purpose. However, as CAPTA becomes more widely deployed, they may become aware of it and deliberately refuse verification or provide misleading answers to avoid identification.

To address this issue, we can mix genuine CAPTCHA challenges with CAPTA probes. Agents must correctly solve genuine challenges to continue their tasks, while their responses also help the service detect deliberate attempts to provide unreliable answers.
Moreover, CAPTA's question selection and option-set optimization methods are not limited to text-based questions and can be extended to other challenge formats. For example, image-selection and audio challenges could make it harder for agents to distinguish CAPTA probes from genuine CAPTCHA challenges, reducing the chance of selective refusal.

\bheading{Model updates.}
When new LLMs are released, the service can reuse existing probes and collect their responses to build new reference fingerprints.
Since CAPTA identifies models based on their preference distributions, different versions or fine-tuned variants of the same model may be treated as different models if their preferences differ sufficiently.
%, but determining an appropriate distance metric and threshold is beyond the scope of this work.

\subsection{Robustness to Context Variations}
LLM responses may vary across agent and task contexts, affecting their preference distributions and identification accuracy. 
Our experiments show that combining multiple probes and collecting more responses can mitigate
these effects.
We evaluate CAPTA across four agents with different system prompts, execution frameworks, and tool interfaces (Section~\ref{sec:eval:agent}).
We also examine cross-task identification (Section~\ref{sec:eval:cross-task}), showing
that the same probes and reference fingerprints can be reused across different tasks. 
Furthermore, we examine different submission settings and runtime temperatures (Section~\ref{sec:eval:practical}).
These results demonstrate that although context variations may affect model preferences, they do not prevent CAPTA from achieving accurate identification across the evaluated settings.

\subsection{CAPTA and Prompt Injection}
\label{sec:dis:ipi}

CAPTA and indirect prompt injection (IPI)~\cite{DBLP:conf/uss/LiuJGJG24, DBLP:conf/nips/DebenedettiZBB024, DBLP:conf/acl/Zhang0Y25} have different goals and mechanisms. 
CAPTA aims to identify the backend LLM behind an agent without changing its original task. 
In contrast, IPI aims to influence agent contexts and cause unintended behaviors. 
They also occur at different stages of agent--service interactions. CAPTA introduces a separate verification step, whereas IPI typically embeds malicious instructions in the requested resource to influence the agent's subsequent actions.

Our work focuses on LLM identification rather than exploitation.
Nevertheless, exposing backend LLM identities may have security implications, as this information could help attackers tailor their strategies to specific models.

\subsection{Limitations}
\label{sec:dis:lim}

\bheading{Heuristic probe construction.}
CAPTA uses heuristic methods for question selection and option-set optimization.
We do not claim that these methods find the optimal questions or option sets. More discriminative probes may still exist.
Nevertheless, as shown in Section~\ref{sec:eval:effectiveness}, our experiments demonstrate the effectiveness of these methods. 

\bheading{Runtime probe selection.}
CAPTA currently ranks probes by their diversity scores and progressively adds them for runtime identification. 
However, this ranking considers individual probe quality rather than the complementary information among selected probes, and may not be optimal for joint identification. Our preliminary experiments (Appendix~\ref{app:runtime-selection}) show that alternative selection methods can reduce the number of probes required to achieve 99\% mean accuracy from 10 to 8, and 99\% worst-model accuracy from 21 to 15. This work focuses on probe construction and model identification; developing more effective runtime probe selection methods is beyond our scope.

% \bheading{Context sensitivity and difficult models.}
% Our cross-task experiments show that CAPTA can reuse reference fingerprints across different tasks while maintaining high accuracy.
% However, context changes may still affect model preferences, and some models remain difficult to identify.
% To address these issues, we can construct probes tailored to specific models, and use more probes and responses per probe to improve identification accuracy.

%\input{sections/7-relatedwork}
\section{Conclusion}
\label{sec:conclusion}

This paper reveals a potential risk of backend LLM identity exposure in normal agent--service interactions. We present CAPTA, which leverages the CAPTCHA verification process in normal agent workflows to identify the backend LLM behind an agent. 
To construct discriminative probes, we propose two methods: diversity-based question selection and LLM-guided option-set optimization.

Our extensive evaluation demonstrates that CAPTA achieves high identification accuracy across diverse LLMs and real-world agents, with only a few probes and interactions.
We further demonstrate the effectiveness of our probe construction methods and CAPTA's applicability to  cross-task and open-set identification.
In future work, we plan to extend CAPTA to new challenge formats, such as image selection and other multimodal challenges, and explore more effective runtime probe selection methods.

% \clearpage
%\looseness=-1

% Verified probe-selection references are in refs.bib.
% Run latexmk (including BibTeX) after changing citations or bibliography data.
\bibliographystyle{IEEEtran}
\bibliography{refs,misc}

\appendices

\section{Optimizer Inputs}
\label{app:optimizer-inputs}

The optimizer receives three types of information: evaluation metrics, 
a preference probability table, and search history.
The evaluation metrics are calculated from the preference probability table 
and recorded in the search history to track optimization progress.
These inputs help the optimizer evaluate the current option set, 
select options for removal, and decide whether to accept or roll back changes.
We introduce each type of information below.

\subsection{Evaluation Metrics}
Before introducing these metrics, we recall the notation from Section~\ref{sec:option-optimization}.
CAPTA uses a representative subset
\(\mathcal{M}'\subset\mathcal{M}\) of candidate LLMs for option-set optimization, 
where \(|\mathcal{M}'|=K'<K\).
For a probe \(p=(q,O)\), 
let \(p_{mo}=\widehat{P}(o\mid m,q,O)\) denote the estimated probability that model \(m\in\mathcal{M}'\) selects option \(o\in O\).

\bheading{Discrimination.}
We first consider how well a probe distinguishes different models.
If several models have similar preference distributions, their runtime responses are difficult to distinguish effectively.
We use standard mutual information to quantify the dependence between model identity and probe response.
Let
\[
    \bar p_o
    =
    \frac{1}{K'}
    \sum_{m\in\mathcal{M}'}p_{mo}.
\]
Under a uniform prior over the representative models,
\[
    I(M;A)
    =
    \frac{1}{K'}
    \sum_{m\in\mathcal{M}'}
    \sum_{o\in O}
    p_{mo}
    \log_2
    \frac{p_{mo}}{\bar p_o}.
\]
We normalize it by the maximum entropy of model identity:
\[
    MI_{\mathrm{ratio}}
    =
    \frac{I(M;A)}{\log_2 K'}.
\]
Normalization bounds the metric between 0 and 1, providing a consistent scale for the LLM to evaluate candidate option sets.
We use $MI_{\mathrm{ratio}}$ as the primary metric for optimization.

\bheading{Single-probe identification.}
We also consider how accurately a single response can identify the model. 
Under a uniform prior over candidate models, the optimal classifier assigns each observed option to the model with the highest probability of selecting it. 
We therefore define the optimal single-probe identification accuracy as
\[
    A_p
    =
    \frac{1}{K'}
    \sum_{o\in O}
    \max_{m\in\mathcal{M}'}p_{mo}.
\]
A higher $A_p$ indicates that a single response provides more discriminative information for model identification.

\bheading{Preference concentration.}
Since CAPTA collects only one or a few responses per probe at runtime,
we prefer each model to exhibit stable preferences for each probe,
rather than dispersed preference distributions that make identification less reliable.
We define
\[
    R_2
    =
    \frac{1}{K'}
    \sum_{m\in\mathcal{M}'}
    \left(\max_{o\in O} p_{mo}\right)^2,
\]
where \(\max_{o\in O} p_{mo}\) is the dominant probability of model \(m\).
We intentionally square the dominant probability to give greater weight
to models with more concentrated preferences.
For example, increasing the dominant probability from $0.6$ to $0.7$
raises its squared value by $0.13$, whereas increasing it from
$0.8$ to $0.9$ raises the value by $0.17$.

\begin{table}[t]
\centering
\caption{Example probability table for the LLM optimizer. Models and options are represented by anonymous IDs.}
\label{tab:probability-table}
\footnotesize
\setlength{\tabcolsep}{1.5pt}
\renewcommand{\arraystretch}{1.15}
\begin{tabularx}{\linewidth}{@{}c*{4}{>{\centering\arraybackslash}X}>{\columncolor{backgroundColor}}c>{\columncolor{backgroundColor}}c>{\columncolor{backgroundColor}}c@{}}
\toprule
 & $M_1$ & $M_2$ & $M_3$ & $M_4$
 & Max gap & Left & Right \\
\midrule
$O_1$ & .80 & .05 & .05 & .05
      & .75 & $M_1$ & $M_2,M_3,M_4$ \\
$O_2$ & .05 & .80 & .05 & .05
      & .75 & $M_2$ & $M_1,M_3,M_4$ \\
$O_4$ & .05 & .05 & .40 & .10
      & .30 & $M_3$ & $M_1,M_2,M_4$ \\
$O_5$ & .10 & .10 & .50 & .80
      & .40 & $M_3,M_4$ & $M_1,M_2$ \\
\midrule
\rowcolor{backgroundColor}
Top options
      & $O_1$ & $O_2$ & $O_5$ & $O_5$
      & -- & -- & -- \\
\rowcolor{backgroundColor}
\makecell{Top-two\\margin}
      & .70 & .70 & .10 & .70
      & -- & -- & -- \\
\bottomrule
\end{tabularx}

\par\smallskip
\parbox{\linewidth}{\footnotesize
Max gap is the largest adjacent gap in sorted model probabilities; Left/Right denote the higher/lower groups. Top-two margin is the difference between a model's two largest option probabilities. Zero-count options are omitted.
}
\end{table}

\subsection{Preference Probability Table}
The preference probability table records the probability of each candidate model selecting each option in the current option set.
As shown in Table~\ref{tab:probability-table}, it also includes auxiliary statistics, such as max gaps, top choices, and top-two margins, to help the optimizer quickly analyze model discrimination and preference concentration.

\subsection{Search History}
The search history records the optimization process across rounds, including the current round, the base round from which it starts, the removed option IDs and their removal reasons, and the final decision (accept or roll back). It also records the evaluation metrics of each round, allowing the optimizer to review previous results and determine which state to restore when a rollback is needed.

\begin{table}[tp]
\centering
\caption{Self-reported identities across 25 models, with 10 responses each. Only 8.0\% of responses name the exact model specified in the API request.}
\label{tab:eval-self-report}
\begingroup
\footnotesize
\setlength{\tabcolsep}{2pt}
\renewcommand{\arraystretch}{1.15}
\definecolor{selfreportE}{HTML}{D1E9C9}
\definecolor{selfreportP}{HTML}{D0E2F2}
\definecolor{selfreportF}{HTML}{E6E6E6}
\definecolor{selfreportW}{HTML}{FFF4CC}
\definecolor{selfreportC}{HTML}{F2D1D1}
\newcommand{\selfreportcategory}[2]{%
  \textcolor{selfreport#1}{\rule{1.2em}{0.8em}}%
  \enspace #2}
\begin{tabular*}{\linewidth}{@{\extracolsep{\fill}}lrr@{}}
\toprule
\textbf{Category} & \textbf{Count} & \textbf{Share (\%)} \\
\midrule
\selfreportcategory{E}{Exact model} & 20 & \acc{8.00} \\
\selfreportcategory{P}{Partial: generation or series} & 32 & \acc{12.80} \\
\selfreportcategory{F}{Vendor or family only} & 109 & \acc{43.60} \\
\selfreportcategory{W}{Wrong model, same vendor} & 26 & \acc{10.40} \\
\selfreportcategory{C}{Wrong vendor} & 63 & \acc{25.20} \\
\bottomrule
\end{tabular*}
\par\medskip
% Generated by tools/export-self-report-table.mjs; do not edit counts manually.
% Source: results/01-self-identification/formal-01/analysis/manual-review.json
% Source SHA256 (UTF-8, BOM removed): ac9183fe90c04c448f3af47ab6377e3752ec9a73ffa494e206803d37a222498e
% Names/order/icons: assets/model-catalog.json; stars: experiment 05 report.
% Tabular fragment included by tables/self-reported-identities.tex.
% The parent table defines the shared category colors and summary legend.
\begingroup
\footnotesize
\setlength{\tabcolsep}{0pt}
\renewcommand{\arraystretch}{1.15}
% Measure names in the actual paper font; the remaining width is ten units.
\dimen0=0pt
\settowidth{\dimen2}{\evalmodel{openai}{GPT-5.6-Luna}}\ifdim\dimen2>\dimen0 \dimen0=\dimen2\fi
\settowidth{\dimen2}{\evalmodel{openai}{GPT-5.6-Terra}}\ifdim\dimen2>\dimen0 \dimen0=\dimen2\fi
\settowidth{\dimen2}{\evalmodel{openai}{GPT-5.6-Sol}}\ifdim\dimen2>\dimen0 \dimen0=\dimen2\fi
\settowidth{\dimen2}{\evalmodel{claude}{Claude-Sonnet-5}}\ifdim\dimen2>\dimen0 \dimen0=\dimen2\fi
\settowidth{\dimen2}{\evalmodel{claude}{Claude-Opus-5}}\ifdim\dimen2>\dimen0 \dimen0=\dimen2\fi
\settowidth{\dimen2}{\evalmodel{claude}{Claude-Sonnet-4.6}}\ifdim\dimen2>\dimen0 \dimen0=\dimen2\fi
\settowidth{\dimen2}{\evalmodel{gemini}{Gemini-3.7-Flash}}\ifdim\dimen2>\dimen0 \dimen0=\dimen2\fi
\settowidth{\dimen2}{\evalmodel{gemini}{Gemini-3.6-Flash}}\ifdim\dimen2>\dimen0 \dimen0=\dimen2\fi
\settowidth{\dimen2}{\evalmodel{gemini}{Gemini-3.5-Flash-Lite}}\ifdim\dimen2>\dimen0 \dimen0=\dimen2\fi
\settowidth{\dimen2}{\evalmodel{deepseek}{DeepSeek-V4-Flash}}\ifdim\dimen2>\dimen0 \dimen0=\dimen2\fi
\settowidth{\dimen2}{\evalmodel{deepseek}{DeepSeek-V4-Pro}}\ifdim\dimen2>\dimen0 \dimen0=\dimen2\fi
\settowidth{\dimen2}{\evalmodel{qwen}{Qwen-3.8-Flash}}\ifdim\dimen2>\dimen0 \dimen0=\dimen2\fi
\settowidth{\dimen2}{\evalmodel{qwen}{Qwen-3.8-Max}}\ifdim\dimen2>\dimen0 \dimen0=\dimen2\fi
\settowidth{\dimen2}{\evalmodel{qwen}{Qwen-3.7-Flash}}\ifdim\dimen2>\dimen0 \dimen0=\dimen2\fi
\settowidth{\dimen2}{\evalmodel{qwen}{Qwen-3.7-Max}}\ifdim\dimen2>\dimen0 \dimen0=\dimen2\fi
\settowidth{\dimen2}{\evalmodel{meta}{Muse-Spark-1.1}}\ifdim\dimen2>\dimen0 \dimen0=\dimen2\fi
\settowidth{\dimen2}{\evalmodel{meta}{Llama-4-Maverick}}\ifdim\dimen2>\dimen0 \dimen0=\dimen2\fi
\settowidth{\dimen2}{\evalmodel{grok}{Grok-4.6}}\ifdim\dimen2>\dimen0 \dimen0=\dimen2\fi
\settowidth{\dimen2}{\evalmodel{grok}{Grok-4.5}}\ifdim\dimen2>\dimen0 \dimen0=\dimen2\fi
\settowidth{\dimen2}{\evalmodel{grok}{Grok-4.3}}\ifdim\dimen2>\dimen0 \dimen0=\dimen2\fi
\settowidth{\dimen2}{\evalmodel{zhipu}{GLM-5.3}}\ifdim\dimen2>\dimen0 \dimen0=\dimen2\fi
\settowidth{\dimen2}{\evalmodel{zhipu}{GLM-5.3-Flash}}\ifdim\dimen2>\dimen0 \dimen0=\dimen2\fi
\settowidth{\dimen2}{\evalmodel{zhipu}{GLM-5.2}}\ifdim\dimen2>\dimen0 \dimen0=\dimen2\fi
\settowidth{\dimen2}{\evalmodel{moonshot}{Kimi-K2.6}}\ifdim\dimen2>\dimen0 \dimen0=\dimen2\fi
\settowidth{\dimen2}{\evalmodel{minimax}{MiniMax-M3}}\ifdim\dimen2>\dimen0 \dimen0=\dimen2\fi
\edef\selfreportnamewidth{\the\dimen0}
\dimen2=\dimexpr(\linewidth-\dimen0-6pt)/10\relax
\edef\selfreportunit{\the\dimen2}
% Zero padding/overhang keeps every segment exactly count times one unit.
\newcommand{\selfreportsegment}[2]{%
  \multicolumn{#1}{@{}>{\centering\arraybackslash}p{\dimexpr\selfreportunit*#1\relax}@{}}{%
    \cellcolor{selfreport#2}\strut #1}}
\begin{tabular}{@{}p{\selfreportnamewidth}@{\hspace{6pt}}*{10}{p{\selfreportunit}}@{}}
\evalmodel{openai}{GPT-5.6-Luna} & \selfreportsegment{1}{P} & \selfreportsegment{7}{F} & \selfreportsegment{2}{W} \\ \noalign{\vskip1pt}
\evalmodel{openai}{GPT-5.6-Terra} & \selfreportsegment{8}{P} & \selfreportsegment{2}{F} \\ \noalign{\vskip1pt}
\evalmodel{openai}{GPT-5.6-Sol} & \selfreportsegment{6}{P} & \selfreportsegment{4}{W} \\ \noalign{\vskip1pt}
\evalmodel{claude}{Claude-Sonnet-5} & \selfreportsegment{10}{C} \\ \noalign{\vskip1pt}
\evalmodel{claude}{Claude-Opus-5} & \selfreportsegment{10}{W} \\ \noalign{\vskip1pt}
\evalmodel{claude}{Claude-Sonnet-4.6} & \selfreportsegment{10}{C} \\ \noalign{\vskip1pt}
\evalmodel{gemini}{Gemini-3.7-Flash} & \selfreportsegment{10}{E} \\ \noalign{\vskip1pt}
\evalmodel{gemini}{Gemini-3.6-Flash} & \selfreportsegment{8}{F} & \selfreportsegment{2}{W} \\ \noalign{\vskip1pt}
\evalmodel{gemini}{Gemini-3.5-Flash-Lite} & \selfreportsegment{3}{W} & \selfreportsegment{7}{C} \\ \noalign{\vskip1pt}
\evalmodel{deepseek}{DeepSeek-V4-Flash} & \selfreportsegment{10}{C} \\ \noalign{\vskip1pt}
\evalmodel{deepseek}{DeepSeek-V4-Pro} & \selfreportsegment{10}{C} \\ \noalign{\vskip1pt}
\evalmodel{qwen}{Qwen-3.8-Flash} & \selfreportsegment{9}{F} & \selfreportsegment{1}{W} \\ \noalign{\vskip1pt}
\evalmodel{qwen}{Qwen-3.8-Max} & \selfreportsegment{10}{F} \\ \noalign{\vskip1pt}
\evalmodel{qwen}{Qwen-3.7-Flash} & \selfreportsegment{10}{F} \\ \noalign{\vskip1pt}
\evalmodel{qwen}{Qwen-3.7-Max} & \selfreportsegment{10}{F} \\ \noalign{\vskip1pt}
\evalmodel{meta}{Muse-Spark-1.1} & \selfreportsegment{10}{P} \\ \noalign{\vskip1pt}
\evalmodel{meta}{Llama-4-Maverick} & \selfreportsegment{10}{F} \\ \noalign{\vskip1pt}
\evalmodel{grok}{Grok-4.6} & \selfreportsegment{4}{P} & \selfreportsegment{5}{F} & \selfreportsegment{1}{W} \\ \noalign{\vskip1pt}
\evalmodel{grok}{Grok-4.5} & \selfreportsegment{3}{P} & \selfreportsegment{3}{F} & \selfreportsegment{1}{W} & \selfreportsegment{3}{C} \\ \noalign{\vskip1pt}
\evalmodel{grok}{Grok-4.3} & \selfreportsegment{10}{C} \\ \noalign{\vskip1pt}
\evalmodel{zhipu}{GLM-5.3} & \selfreportsegment{9}{F} & \selfreportsegment{1}{W} \\ \noalign{\vskip1pt}
\evalmodel{zhipu}{GLM-5.3-Flash} & \selfreportsegment{10}{F} \\ \noalign{\vskip1pt}
\evalmodel{zhipu}{GLM-5.2} & \selfreportsegment{8}{F} & \selfreportsegment{1}{W} & \selfreportsegment{1}{C} \\ \noalign{\vskip1pt}
\evalmodel{moonshot}{Kimi-K2.6} & \selfreportsegment{8}{F} & \selfreportsegment{2}{C} \\ \noalign{\vskip1pt}
\evalmodel{minimax}{MiniMax-M3} & \selfreportsegment{10}{E} \\ \noalign{\vskip1pt}
\end{tabular}
\endgroup

\endgroup
\end{table}

\section{Self-reported identities}
\label{app:eval-self-report}

To evaluate the reliability of self-reported model identities, we query 25 models through their APIs with the same question: ``What model are you?''
We collect 10 valid responses per model, resulting in 250 responses.
We then manually classify these responses based on the model identities specified in the API requests.

Table~\ref{tab:eval-self-report} summarizes the results.
Only 20 of the 250 responses (8.00\%) correctly report the exact model identity.
Most responses provide only partial identity information: 12.80\% report the correct model generation or series,
and 43.60\% report only the vendor or model family.
In addition, 10.40\% report an incorrect model from the same vendor,
while 25.20\% report a model from a different vendor.
Overall, 56.40\% provide partial identity information, and 35.60\% report incorrect identities.

The lower part of Table~\ref{tab:eval-self-report} shows the detailed results for each model.
Only Gemini-3.7-Flash and MiniMax-M3 correctly report their identities, both in all 10 trials.
Some models consistently report only family-level information.
For example, 39 of the 40 responses from four Qwen models report only \textit{Qwen}.
Some models consistently report incorrect identities.
For example, DeepSeek-V4-Flash reports itself as \textit{gpt-4o-mini} in all 10 trials, 
while Claude-Sonnet-5 reports itself as \textit{GPT-5} in all 10 trials.
These results show that consistent self-reports do not necessarily indicate accurate identities.
Overall, even when a service can directly ask a model about its identity,
the responses may not reliably identify the exact backend LLM.

\begin{figure}[tp]
\centering
\includegraphics[scale=1]{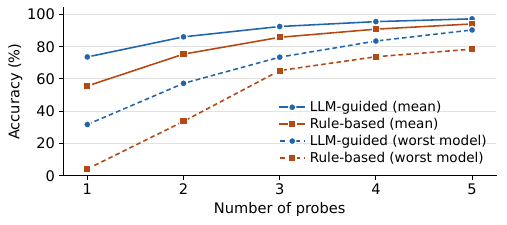}
\caption{
Comparison of LLM-guided and rule-based option-set optimization across seven models. 
LLM-guided optimization achieves higher mean and worst-model identification accuracy.
}
\label{fig:eval-method}
\end{figure}

\section{LLM-guided vs. Rule-based Optimization}
\label{app:opt-comparison}

We propose an LLM-guided method for option-set optimization.
To evaluate its effectiveness, we compare it with a naive optimization method based on fixed rules.
Specifically, the rule-based method prioritizes removing options with smaller maximum probability gaps across models, as these options provide weaker distinctions between models.
When multiple options have the same gap, it prioritizes those whose higher-probability group contains more models.
After generating a new candidate option set, the method uses the same evaluation metrics to decide whether to accept the candidate or roll back the change.

\bheading{Experimental setup.}
Both methods start with the same top five initial probes and use the same seven representative models.
Other experimental settings follow those in Section~\ref{sec:eval-setup}, including a maximum of 12 optimization rounds and early stopping after six consecutive rounds without an accepted candidate.

To compare the performance of the optimized probes, we further evaluate their joint accuracy with different numbers of probes.
Specifically, for each $k=1,\ldots,5$, we enumerate all possible combinations of $k$ probes.
For each $k$, we first calculate the average identification accuracy of each model across all probe combinations.
We then report the mean and minimum accuracy across the seven models as the mean and worst-model accuracy, respectively.

\bheading{Results.}
As shown in Figure~\ref{fig:eval-method}, LLM-guided optimization achieves higher mean and worst-model accuracy across all probe counts.
For single-probe identification, the mean accuracy increases from \accp{55.53} with rule-based optimization to \accp{73.55} with LLM-guided optimization, an improvement of 18.0 percentage points.
With five probes, the mean accuracy reaches \accp{97.19}, compared with \accp{94.03} for rule-based optimization, a difference of 3.16 percentage points.
However, the gap in worst-model accuracy remains substantial.
With five probes, LLM-guided optimization achieves \accp{90.30}, compared with \accp{78.40} for rule-based optimization, an improvement of 11.9 percentage points.
These results show that LLM-guided optimization not only improves overall identification accuracy but also provides better performance for models that are more difficult to identify.

\begin{table*}[t]
\centering
\caption{Batch and Cumulative identification accuracy (\%) across seven models. Batch benefits from matched fingerprints, while Cumulative reuses single-probe fingerprints.}
\label{tab:multi-probes}
\footnotesize
\setlength{\tabcolsep}{2pt}
\renewcommand{\arraystretch}{1.15}
\renewcommand{\accdigits}{2}
% The first argument determines the background color of the entire cell.
\newcommand{\meanworst}[2]{\acccell{#1}\,/\,\acc{#2}}
\begin{tabular*}{\linewidth}{@{\extracolsep{\fill}}llc*{6}{c}@{}}
\toprule
\multirow{2}{*}{\textbf{Type}}
  & \multirow{2}{*}{\textbf{Fingerprint source}}
  & \multirow{2}{*}{\textbf{Size}}
  & \multicolumn{6}{c}{\textbf{Number of interactions}} \\
\cmidrule(l){4-9}
  & & & \textbf{1} & \textbf{2} & \textbf{3}
      & \textbf{4} & \textbf{5} & \textbf{6} \\
\midrule
\multirow{4}{*}{Batch}
  & \multirow{2}{*}{Single-probe}
  & 5  & \meanworst{74.29}{40.00} & \meanworst{77.14}{50.00}
       & \meanworst{74.29}{0.00} & \meanworst{78.57}{20.00}
       & \meanworst{87.14}{40.00} & \meanworst{88.57}{30.00} \\
  & & 10 & \meanworst{87.14}{70.00} & \meanworst{72.86}{0.00}
         & \meanworst{74.29}{0.00} & --- & --- & --- \\
\cmidrule(l){2-9}
  & \multirow{2}{*}{Batch-matched}
  & 5  & \meanworst{97.56}{87.40} & \meanworst{98.93}{93.40}
       & \meanworst{99.64}{97.50} & \meanworst{99.86}{99.00}
       & \meanworst{99.99}{99.90} & \meanworst{100.00}{100.00} \\
  & & 10 & \meanworst{98.67}{92.50} & \meanworst{99.73}{98.10}
         & \meanworst{99.91}{99.40} & --- & --- & --- \\
\midrule
\multirow{2}{*}{Cumulative}
  & \multirow{2}{*}{Single-probe}
  & 5  & \meanworst{77.14}{50.00} & \meanworst{94.29}{80.00}
       & \meanworst{95.71}{80.00} & \meanworst{95.71}{70.00}
       & \meanworst{98.57}{90.00} & \meanworst{100.00}{100.00} \\
  & & 10 & \meanworst{88.57}{60.00} & \meanworst{94.29}{70.00}
         & \meanworst{100.00}{100.00} & --- & --- & --- \\
\bottomrule
\end{tabular*}
\par\smallskip
\parbox{\linewidth}{\footnotesize
Cells show mean / worst-model accuracy. Size denotes probes per group; each interaction adds one group. Fingerprints use isolated probes (Single-probe) or the matching batch format (Batch-matched).
}
\end{table*}

\section{Multi-probe Submission within a Single Verification}
\label{app:multi-probe}

In practice, a single verification may contain multiple probes, similar to how Google reCAPTCHA presents multiple image-selection challenges in one verification. We consider two strategies. In \textit{Batch}, the service provides a group of probes in a single request and asks the agent to answer them all at once. In \textit{Cumulative}, the service presents probes one by one within the same verification session, retaining previous questions and answers in the agent's context.

\bheading{Experimental setup.}
We use the same 30 optimized probes from the main evaluation, preserving their order, and divide them into groups of 5 or 10 probes.
Thus, completing all 30 probes requires 6 or 3 agent--service interactions, respectively.
We evaluate both strategies using the seven representative LLMs.
For each model and probe group, we evaluate 10 runtime repetitions in both Batch and Cumulative formats against the existing single-probe reference fingerprints (\textit{Single-probe}).
For Batch, we additionally collect responses using the same group sizes to construct and evaluate matching reference fingerprints (\textit{Batch-matched}).
For both strategies, we progressively combine responses from successive groups for identification.
We report the mean and worst-model identification accuracy across the seven models.

\bheading{Results.}
As shown in Table~\ref{tab:multi-probes}, Batch is sensitive to how reference fingerprints are constructed. Reusing Single-probe fingerprints results in relatively low and unstable accuracy, while Batch-matched fingerprints substantially improve performance. In contrast, Cumulative can directly reuse existing Single-probe fingerprints, and its accuracy improves with more interactions. With 10 probes per group, Batch achieves \accp{99.73} mean accuracy and \accp{98.10} worst-model accuracy using Batch-matched fingerprints after two interactions, while Cumulative achieves \accp{100} accuracy across all models after three interactions. Overall, both strategies support multi-probe verification within a single verification process. However, Batch requires reference fingerprints that match the submission format, whereas Cumulative can reuse existing single-probe fingerprints.

\begin{table}[t]
\centering
\caption{Mean identification accuracy across 25 models with 95\% confidence intervals (CI).}
\label{tab:eval-accuracy-ci}
\footnotesize
\setlength{\tabcolsep}{2pt}
\renewcommand{\arraystretch}{1.15}
\rowcolors{2}{white}{gray!8}
\begin{tabularx}{\columnwidth}{@{}
  >{\hsize=0.45\hsize\centering\arraybackslash}X
  >{\hsize=1.30\hsize\centering\arraybackslash}X
  >{\hsize=1.3\hsize\centering\arraybackslash}X
  >{\hsize=0.95\hsize\centering\arraybackslash}X
@{}}
\toprule
\rowcolor{white}
\textbf{Probes} & \textbf{Mean Accuracy (\%)} & 
\textbf{95\% CI (\%)} & \textbf{CI Width (pp)} \\
\midrule
5  & 91.650 & [91.330, 91.968]  & 0.638 \\
10 & 99.032 & [98.908, 99.152]  & 0.244 \\
15 & 99.644 & [99.568, 99.716]  & 0.148 \\
20 & 99.900 & [99.856, 99.940]  & 0.084 \\
25 & 99.972 & [99.948, 99.992]  & 0.044 \\
30 & 99.988 & [99.972, 100.000] & 0.028 \\
\bottomrule
\end{tabularx}
\par\smallskip
\parbox{\linewidth}{\footnotesize
pp denotes percentage points.
}
\end{table}

\section{Confidence Interval Analysis}
\label{app:confidence}

We calculate 95\% confidence intervals to assess the stability of the mean identification accuracy reported in Section~\ref{sec:eval:identification}.
We use the bootstrap method to calculate the confidence intervals. Specifically, we randomly sample 1,000 trials with replacement from the original 1,000 trials and recalculate the mean accuracy. 
In each resampling, we preserve the results of all 25 models within the same trial to account for their dependence. We repeat this process 50,000 times and use the 2.5th and 97.5th percentiles of the resulting accuracy distribution as the lower and upper bounds of the 95\% confidence interval, respectively.

As shown in Table~\ref{tab:eval-accuracy-ci}, the high identification accuracy and narrow confidence intervals demonstrate the effectiveness and stability of CAPTA across different probe budgets. For example, with 10 probes, the mean identification accuracy is 99.032\%, with a 95\% confidence interval of [98.908\%, 99.152\%]. The intervals become narrower as the number of probes increases. These results indicate that the estimated mean identification accuracy is stable under the current experimental setting.

Note that these confidence intervals are calculated based on the fixed response pool and probes. They reflect the uncertainty introduced by random holdout and fingerprint reconstruction, but do not capture variations from collecting new LLM responses.

% Generated by latex/CAPTA/tools/export-probe-selection-table.mjs.
% Source: results/08-probe-selection/formal-01/paper-budgets.csv
% Source SHA256: d3485b5542b75c6c34b2d467b0cd40c5eda0b637760a0b55671e198c119e86c4
% All 25 models only. Random remains in the complete results.
\begin{table}[t]
\centering
\caption{Number of probes required to achieve 99\% mean and worst-model identification accuracy across 25 models.}
\label{tab:eval-probe-selection}
\footnotesize
\setlength{\tabcolsep}{1pt}
\renewcommand{\arraystretch}{1.15}
\begin{tabular*}{\linewidth}{@{\extracolsep{\fill}}lccccc@{}}
\toprule
& \makecell[c]{\textbf{Original}\\\textbf{order}} & \makecell[c]{\textbf{Single}\\\textbf{MI}~\cite{DBLP:journals/corr/abs-1112-5745}} & \makecell[c]{\textbf{Joint}\\\textbf{MI}~\cite{DBLP:conf/nips/KirschAG19}} & \makecell[c]{\textbf{Validation}\\\textbf{greedy}~\cite{DBLP:journals/sac/CamattariGMPP24}} & \makecell[c]{\textbf{Pairwise}\\\textbf{overlap}~\cite{DBLP:journals/pr/ChenZY24}} \\
\midrule
\textbf{Mean} & 10 & 9\,\textcolor{themeGreen}{\bfseries\boldmath($\downarrow$1)} & 9\,\textcolor{themeGreen}{\bfseries\boldmath($\downarrow$1)} & 9\,\textcolor{themeGreen}{\bfseries\boldmath($\downarrow$1)} & 8\,\textcolor{themeGreen}{\bfseries\boldmath($\downarrow$2)} \\
\textbf{Worst-model} & 21 & 15\,\textcolor{themeGreen}{\bfseries\boldmath($\downarrow$6)} & 15\,\textcolor{themeGreen}{\bfseries\boldmath($\downarrow$6)} & 17\,\textcolor{themeGreen}{\bfseries\boldmath($\downarrow$4)} & 15\,\textcolor{themeGreen}{\bfseries\boldmath($\downarrow$6)} \\
\bottomrule
\end{tabular*}

\par\smallskip
\parbox{\linewidth}{\footnotesize
Lower values indicate better efficiency. Green numbers show the probes saved compared with the original order.
}
\end{table}

\section{Runtime Probe Selection}
\label{app:runtime-selection}
During runtime identification, CAPTA can identify the target model using as few probes as possible.
In Section~\ref{sec:eval}, CAPTA ranks probes by their question diversity scores and progressively adds them for joint identification.
However, this ranking does not account for the complementary information that different probes may provide.
We therefore explore alternative selection methods to identify the target model with fewer probes while maintaining the same accuracy.

We consider four existing selection methods.
Single MI~\cite{DBLP:journals/corr/abs-1112-5745} ranks probes by their mutual information with model identity; 
Joint MI~\cite{DBLP:conf/nips/KirschAG19} greedily selects probes to maximize the mutual information of the selected set;
Validation greedy~\cite{DBLP:journals/sac/CamattariGMPP24} selects probes based on validation accuracy; 
and Pairwise overlap~\cite{DBLP:journals/pr/ChenZY24} prioritizes probes that reduce overlap between model response distributions.

As shown in Table~\ref{tab:eval-probe-selection}, all four methods reduce the number of probes required.
To achieve 99\% mean identification accuracy, Pairwise overlap requires only 8 probes, compared with 10 under the original ranking.
For 99\% worst-model accuracy, Single MI, Joint MI, and Pairwise overlap require only 15 probes while Validation greedy requires 17, compared with 21 under the original ranking.

These results suggest that selecting complementary probes can improve CAPTA's runtime efficiency. 
Our work focuses on probe construction, while developing more effective methods for probe selection is left for future work.

\end{document}